\newcommand{\subparagraph}{}

\documentclass[journal]{IEEEtran}
\usepackage{cite}

\ifCLASSINFOpdf
	\usepackage{graphicx}
	\graphicspath{{./images/}}
	\DeclareGraphicsExtensions{.pdf,.eps}
\else
\fi
\usepackage{amsmath}
\usepackage{array}

\usepackage{soul}
\usepackage{color}
\usepackage{bm}
\usepackage{multirow}
\usepackage{threeparttable}
\usepackage{hyperref}

\usepackage{titlesec}

\usepackage{fancyhdr}
\usepackage{xcolor}

\fancypagestyle{firstpage}{
\fancyhf{} 
\fancyhead[L]{}
\fancyhead[C]{\fontsize{8}{9.6}\selectfont{This is the author's version of an article that has been published in IEEE TMI. Changes were made to this version by the publisher prior to publication. The final version of record is available at {\color{blue}\url{http://dx.doi.org/10.1109/TMI.2019.2923601}}}}
\fancyhead[R]{}
\fancyfoot[L]{}
\fancyfoot[C]{\fontsize{8}{9.6}\selectfont{Copyright (c) 2019 IEEE. Personal use is permitted. For any other purposes, permission must be obtained from the IEEE by emailing  pubs-permissions@ieee.org.}}
\fancyfoot[R]{}

}

\begin{document}
%
\title{Co-Learning Feature Fusion Maps from PET-CT Images of Lung Cancer}
%
%
%

\author{Ashnil~Kumar,~\IEEEmembership{Member,~IEEE,}
        Michael~Fulham,
       Dagan~Feng,~\IEEEmembership{Fellow, IEEE,} and~Jinman~Kim,~\IEEEmembership{Member,~IEEE}
\thanks{A. Kumar, D. Feng, and J. Kim are with the School of Computer Science, University of Sydney, Australia.}
\thanks{M. Fulham is with the Department of Molecular Imaging, Royal Prince Alfred Hospital, Australia, and also with the Sydney Medical School, University of Sydney, Australia.}
\thanks{This work was supported in part by ARC grants.}%
\thanks{This paper has supplementary downloadable material available at \url{http://ieeexplore.ieee.org}.}%
\thanks{Copyright~\copyright~2019 IEEE. Personal use of this material is permitted. However, permission to use this material for any other purposes must be obtained from the IEEE by sending a request to \url{pubs-permissions@ieee.org}.}}



%



\maketitle
\thispagestyle{firstpage}

\begin{abstract}
The analysis of multi-modality positron emission tomography and computed tomography (PET-CT) images for computer aided diagnosis applications (e.g., detection and segmentation) requires combining the sensitivity of PET to detect abnormal regions with anatomical localization from CT. Current methods for PET-CT image analysis either process the modalities separately or fuse information from each modality based on knowledge about the image analysis task. These methods generally do not consider the spatially varying visual characteristics that encode different information across the different modalities, which have different priorities at different locations. For example, a high abnormal PET uptake in the lungs is more meaningful for tumor detection than physiological PET uptake in the heart. Our aim is to improve fusion of the complementary information in multi-modality PET-CT with a new supervised convolutional neural network (CNN) that learns to fuse complementary information for multi-modality medical image analysis. Our CNN first encodes modality-specific features and then uses them to derive a spatially varying fusion map that quantifies the relative importance of each modality's features across different spatial locations. These fusion maps are then multiplied with the modality-specific feature maps to obtain a representation of the complementary multi-modality information at different locations, which can then be used for image analysis. We evaluated the ability of our CNN to detect and segment multiple regions (lungs, mediastinum, tumors) with different fusion requirements using a dataset of PET-CT images of lung cancer. We compared our method to baseline techniques for multi-modality image fusion (fused inputs (FS), multi-branch (MB) techniques, and multi-channel (MC) techniques) and segmentation. Our findings show that our CNN had a significantly higher foreground detection accuracy (99.29\%, $p < 0.05$) than the fusion baselines (FS: 99.00\%, MB: 99.08\%, TC: 98.92\%) and a significantly higher Dice score (63.85\%) than recent PET-CT tumor segmentation methods.
\end{abstract}


%
\IEEEpeerreviewmaketitle

\section{Introduction}
\label{sec:intro}
\IEEEPARstart{M}{edical} imaging is a cornerstone of modern healthcare providing unique diagnostic, and increasingly therapeutic, capabilities that affect patient care. The range of medical imaging modalities is wide but in essence they provide anatomical and functional information about structure and physiopathology. The multi-modality $^{18}$F-Fluorodeoxyglucose (FDG) positron emission tomography and computed tomography (PET-CT) scanner is regarded as the imaging device of choice for the diagnosis, staging, and assessment of treatment response in many cancers~\cite{kligerman2009_petctlung}. PET-CT combines the sensitivity of PET to detect regions of abnormal function and anatomical localization from CT~\cite{blodgett2007_petct}. With PET, sites of disease usually display greater FDG uptake (glucose metabolism) than normal structures. The spatial extent of the disease within a particular structure, however, cannot be accurately determined due to the inherent lower resolution of PET when compared to CT and MR imaging, tumor heterogeneity, and the partial volume effect~\cite{wahl2009_percistpetct}. CT provides the anatomical localization of sites of abnormal FDG uptake in PET and so adds precision to image interpretation~\cite{bagci2013_multimodseg}. One example clinical domain that has benefited greatly from PET-CT imaging is the evaluation of non-small cell lung cancer (NSCLC), the most common type of lung cancer. In NSCLC, the extent of the disease at diagnosis is the most important determinant of patient outcome: whether it is restricted to the lung parenchyma, involved the lung plus regional lymph nodes at the pulmonary hilum and/or mediastinum, extends into the mediastinum or chest wall directly, or has spread beyond the thorax. PET-CT is able to detect sites of disease where there are no abnormalities in the underlying structure on CT, hence its value in patient management~\cite{detterbeck2009_cancerstage,edge2010_ajcc,edge2010_ajcctnm,tatci2015_petctmediasitnum}.

The role of PET-CT in cancer care has provoked extensive research into methods to detect, classify, and retrieve PET-CT images~\cite{teramoto2016_petctpulmnod,bi2017_petctclass,xu2018_petctbone,milletari2016_vnet,zhong2018_petctunet,zhao2015_petctpulmnodsvm,lartizien2014_cadpetct,song2014_lesiondetectseppetct,song2013_petcttumsegsepgraph,han2011_ipmipetctseg,bradshaw2018_vgg3chanpetct,song2012_tumlymhpetct,bi2014_petctmultistage,kumar2014_media}. These methods can be separated into those that: (i) process each modality separately and then combine the modality-specific features~\cite{teramoto2016_petctpulmnod,bi2017_petctclass,xu2018_petctbone,milletari2016_vnet,zhong2018_petctunet,zhao2015_petctpulmnodsvm,lartizien2014_cadpetct,song2014_lesiondetectseppetct,song2013_petcttumsegsepgraph,han2011_ipmipetctseg}, and (ii) combine or fuse complementary features from each modality~\cite{bradshaw2018_vgg3chanpetct,song2012_tumlymhpetct,bi2014_petctmultistage,kumar2014_media}. Methods that process each modality separately are inherently limited  when the intent is to consider both the function and anatomical extent of disease. For example, in a chest CT depicting a lung tumor that is causing collapse in adjacent lung tissue, both the tumor and the collapse can appear identical. Similarly, some areas of high FDG uptake on  PET images may be linked to normal physiological uptake, such as in the heart, and these regions need to be filtered out based upon knowledge about anatomical characteristics from CT to differentiate them from abnormal PET regions~\cite{boellaard2015_petctguidelines,kroiss2013_petctuptake,rosenbaum2006_falsepositiveuptake,blodgett2005_petctheadneck}. In contrast, methods that fuse information from the two modalities often use \emph{a priori} knowledge about characteristics of the different modalities to `prioritize' information from one of the two modalities for different tasks. Alternatively, they may fuse information using a representation that models relationships between the two modalities. The fusion is particularly necessary in cases where different imaging modalities identify different attributes of the same region of interest (ROI), with no one modality capturing the entire ROI~\cite{bird2015_multimodgtv}. Bagci et al.~\cite{bagci2013_multimodseg} proposed a method to simultaneously delineate ROIs in PET, PET-CT, PET-MR imaging, and fused MR-PET-CT images using a random walk segmentation algorithm with an automated similarity-based seed selection process. Zhao et al.~\cite{zhao2015_petctpulmnodsvm} combined dynamic thresholding, watershed segmentation, and support vector machine (SVM) classification to classify solitary pulmonary nodules on the basis of CT texture features and PET metabolic features. Similarly, Lartizien et al.~\cite{lartizien2014_cadpetct} used texture feature selection and SVM classification for staging of lymphoma patients. Y. Song et al.~\cite{song2014_lesiondetectseppetct}, Q. Song et al.~\cite{song2013_petcttumsegsepgraph}, and Ju et al.~\cite{ju2015_randwalkcutsegpetct} used the context of PET and CT regions to characterize tumors with spatial and visual consistency. Han et al.~\cite{han2011_ipmipetctseg} segmented tumors from PET-CT images, formulating the problem as a Markov Random Field with modality-specific energy terms for PET and CT characteristics. In our prior work~\cite{song2012_tumlymhpetct,bi2014_petctmultistage}, we used multi-stage discriminative models to classify ROIs in the thoracic PET-CT images and in full-body lymphoma studies. In our PET-CT retrieval research~\cite{kumar2014_media,kumar2014_embc}, we have also derived a graph-based model that attempts to bridge the semantic gap by modeling the spatial characteristics that are important for lung cancer staging~\cite{detterbeck2009_cancerstage}. These methods are highly dependent upon an external predefined specification of the relationship between the features from both modalities. Hence, the ability to derive an application-specific fusion would reduce this dependency.

For multi-modality medical imaging, fusion is necessary for computer aided diagnosis tasks such as image visualization, lesion detection, lesion segmentation, and disease classification. For example, Li et al.~\cite{li2018_fusiondenoiseenhance} designed a fusion technique for denoising and enhancing the visual details in multi-modality medical images, while Tong et al.~\cite{tong2017_mmclassalz} fused information from MR and PET images with clinical and genetic biomarkers to predict Alzheimer's disease. Current image fusion strategies in the general (non-medical) domain derive spatially varying fusion weights from the local visual characteristics of the different image data~\cite{li2017_fusionreview}. Features such as pixel variance, contrast, and color saturation are used to derive task-specific fusion ratios for different regions of interest (ROIs) within the images~\cite{kumar2009_variationfusion,shen2012_multifusion}. These fusion methods can thus adapt to and prioritize different content at different locations in the images according to the underlying image features that are relevant to the different images being analyzed. This results in the capacity to enhance specific information from different image data for different tasks. In the case of cancer, a disease that can spread throughout the body, a spatially varying fusion may enhance analysis of the multi-modality medical image data~\cite{hatt2011_petcttumheterogseg}. This may enable greater scrutiny of heterogeneous tumors, better investigation of tumours at and across different tissue boundaries (e.g., tumor invasion of adjacent anatomical structures)~\cite{nagamachi2013_petmrinvasion}, and through integration into visualisation pipelines improve clinicians' interpretation of a patient’s data (e.g., for disease staging)~\cite{boellaard2009_eanmpetctguide}.

Our hypothesis is that the derivation of an appropriate spatially varying fusion of multi-modality image data should be learned from the underlying visual features of the individual images, as this will enable a better integration of the complementary information within each modality. The state-of-the-art in feature learning, selection, and extraction are deep learning methods~\cite{bengio2013_replearn,lecun2015_DLnature}. Convolutional neural networks (CNNs)~\cite{krizhevsky2012_alexnet} are deep learning methods for object detection, classification, and analysis of image data. CNNs have achieved better results than non-deep learning methods in many benchmark tasks, e.g., the ImageNet Large Scale Visual Recognition Challenge~\cite{russakovsky2015_imagenet}. This dominance relates to the ability of CNNs to implicitly learn image features that are `meaningful' for a given task directly from the image data. 

Initial research in medical image analysis used CNN approaches that `transferred' features learned from a non-medical domain and tuned them to a specific medical task~\cite{shin2016_cnntransfer,tajbakhsh2016_cnnfinetune}, e.g., classification of the modality of the medical images depicted in research literature~\cite{kumar2017_jbhi}, and the localization of planes in fetal ultrasound images~\cite{chen2015_usloc}. Later studies designed new CNNs for specific clinical challenges, such as the classification of interstitial lung disease in CT images~\cite{anthimopoulos2016_cnnild}. Transfer learning approaches have also been used for multi-modality image classification~\cite{bi2017_petctclass,bradshaw2018_vgg3chanpetct}. Bi et al.~\cite{bi2017_petctclass} used domain-transferred CNNs to extract PET features for PET-CT lymphoma classification. Bradshaw et al.~\cite{bradshaw2018_vgg3chanpetct} fine-tuned a CNN (pre-trained on ImageNet data) for PET-CT images in a multi-channel input approach, using a CT slice and two maximum intensity projections of the PET data as the inputs. 

CNNs have also been specifically trained for a number of multi-modality medical image analysis applications. One  area of focus are brain MR images obtained with different sequences that are often treated as multi-modality images with the reasoning that the different MR images showed different aspects of the same anatomical structure~\cite{zhang2015_neuromm,tseng2017_crossmodconv,vantulder2017_sharedcrossmod}. Zhang et al.~\cite{zhang2015_neuromm} designed a CNN-based segmentation approach for brain MR images based upon this reasoning. Similarly, Tseng et al.~\cite{tseng2017_crossmodconv} segmented ROIs with complementary features that were learned via a convolution across the different MR images. Van Tulder and de Bruijne~\cite{vantulder2017_sharedcrossmod} used an unsupervised approach to learn a shared data representation of MR images, which acted as a robust feature descriptor for classification applications. In the wider, multi-modality image domain, Liu et al.~\cite{liu2018_mrctattencorr} used a convolutional autoencoder to detect air, bone, and soft tissue for attenuation correction in PET-MR images. Teramoto et al.~\cite{teramoto2016_petctpulmnod} used a CNN as a second stage classifier to determine if candidate lung nodules in PET-CT were false positives. Xu et al.~\cite{xu2018_petctbone} cascaded two V-Nets~\cite{milletari2016_vnet} to detect bone lesions, using CT alone as the input to the first V-Net and a pre-fused PET-CT image for the second. Zhao et al.~\cite{zhao2018_tumorpetctseg} also used V-Nets in a multi-branch paradigm for lung tumor segmentation. Similarly, Zhong et al.~\cite{zhong2018_petctunet} trained one U-Net~\cite{ronneberger2015_unet} for PET and one for CT, combining the results using a graph cut algorithm. Li et al.~\cite{li2019_dlpetctseg} reported a variational model that integrated PET pixel intensity with a CNN-derived CT probability map to segment lung tumors. In general,  CNNs have been applied to multi-modality image data as feature extractors and classifiers without consideration of how the features from each modality were combined, relying either on pre-fusion of the input data or independent processing of each input modality. In addition, recent research on CNN-based PET-CT lung tumor segmentation~\cite{zhao2018_tumorpetctseg,zhong2018_petctunet,li2019_dlpetctseg} learned from image patches centered around the tumor and did not consider variation of PET and CT image features for tumors occurring in different anatomical locations. 

Our aim was to improve fusion of the complementary information in multi-modality images for automatic medical image analysis. In particular, we focus on image data that depict disease across multiple anatomical locations. We present a new CNN that learns to fuse complementary anatomical and functional data from PET-CT images in a spatially varying manner. The novelty of our CNN is its ability to produce a fusion map that explicitly quantifies the fusion weights for the features in each modality. This is in contrast to CNNs that use multi-channel inputs~\cite{bradshaw2018_vgg3chanpetct,zhang2015_neuromm}, where modalities are implicitly fused, or modality-specific encoder branches~\cite{teramoto2016_petctpulmnod,vantulder2017_sharedcrossmod,zhong2018_petctunet,zhong2019_simultaneouscosegpetct}, where the modality-specific features are concatenated at a later stage. Our co-learning CNN is intended as a general approach for integrating PET and CT information, with components that can be leveraged and optimized for a number of different medical image analysis tasks, such as visualization, classification, and segmentation. To demonstrate the efficacy of our CNN, we conducted experimental comparisons with baseline fusion and tumor segmentation methods on PET-CT lung cancer images.

\section{Methods}
\label{sec:methods}

\subsection{Materials}
\label{sec:methods:materials}

Our dataset comprised 50 FDG PET-CT scans of patients with biopsy-proven NSCLC. Since our intention was to analyze the thorax, our imaging specialist chose representative cases that included patients with solitary lung primary tumors with and without hilar (Stage II) and mediastinal (Stage III) nodal involvement and those where the tumour involved the mediastinum and the chest wall. These cases were chosen from the imaging archive of the Department of Molecular Imaging at the Royal Prince Alfred Hospital, Sydney, Australia over consecutive cases in a three month period. All studies were acquired on a Biograph 128-slice mCT (PET-CT scanner; Siemens Healthineers, Hoffman Estates, Il, USA). The mCT is a high-resolution tomograph with high-definition reconstruction, time-of-flight and flow motion characteristics. Each study comprised one CT volume and one PET volume: the CT resolution was $512 \times 512$ pixels at 0.98mm $\times$ 0.98mm, the PET resolution was 200 $\times$ 200 pixels at 4.07mm $\times$ 4.07mm, with a slice thickness and an interslice distance of 3mm. Both volumes were reconstructed with the same number of slices. Studies contained between 1 to 7 tumors (inclusive) in the thorax. The tumor locations included the different lung lobes, the mediastinum, and hilar nodes. All data were de-identified. 

The images were rescaled to a resolution of $256 \times 256$ pixels (x-y axes); rescaling the PET and CT volumes so that they share the same coordinate space is a standard process for analysis of PET-CT data~\cite{zhong2019_simultaneouscosegpetct,han2011_ipmipetctseg,li2019_dlpetctseg,zhao2018_tumorpetctseg}. The PET images were normalized by a transformation to standard uptake values (SUVs). The SUV is a semi-quantitative value of the degree of FDG taken up by the sites of tumor relative to isotope dose and patient mass~\cite{thie2004_suv}. The thorax subvolume (set of 2D thorax slices) of each study were manually identified (mean: 86.5 slices, standard deviation: 6.8 slices). To ensure a balanced class distribution for CNN training, we used the ground truth to select only the axial thorax slices containing all ROIs (lungs, mediastinum, tumors), producing a final dataset of 855 PET-CT slice pairs (855 CT, 855 PET). This sampling  is a standard strategy for learning from imbalanced data~\cite{he2008_unbalancedlearning}.

The ground truth was derived from the diagnostic imaging report which detailed the locations of the primary tumor and any involved thoracic lymph nodes. All reports were done by a single, experienced imaging specialist who has read over 80,000 PET and PET-CT scans. We used the report findings to drive a semi-automatic process for ROI labeling. We applied a commonly used adaptive thresholding algorithm~\cite{hu2001_lungseg} to extract the lung ground truth from the CT volume. Similarly, we used connected thresholding to coarsely determine the mediastinum. We extracted the tumor ground truth using 40\% peak SUV connected thresholding to detect the `hot spots’ identified in the diagnostic reports, which is a general cutoff that is widely used when examining tumors in PET images~\cite{bradley2004_suvseg,hong2007_correlationpetsuvuptake,pak2018_prognosticpetsuv,morand2018_suvmax}. Minor manual adjustments were performed to facilitate extraction of the ground truth ROIs (e.g., preventing the left and right lung fields from being joined together with the edge of the mediastinum).

We randomly divided the 50 PET-CT studies into 5 distinct training and test sets for use in a 5-fold cross validation evaluation protocol (see Section~\ref{sec:methods:eval}). Each training set comprised the slices from 40 studies and its associated test set comprised slices from the 10 other studies. A step-by-step description of our dataset creation process is provided in the Supplementary Materials (Section~SIV).

\subsection{Architecture Design}
\label{sec:methods:arch}

\begin{figure}[!t]
	\centering
    \includegraphics[width=\columnwidth,keepaspectratio]{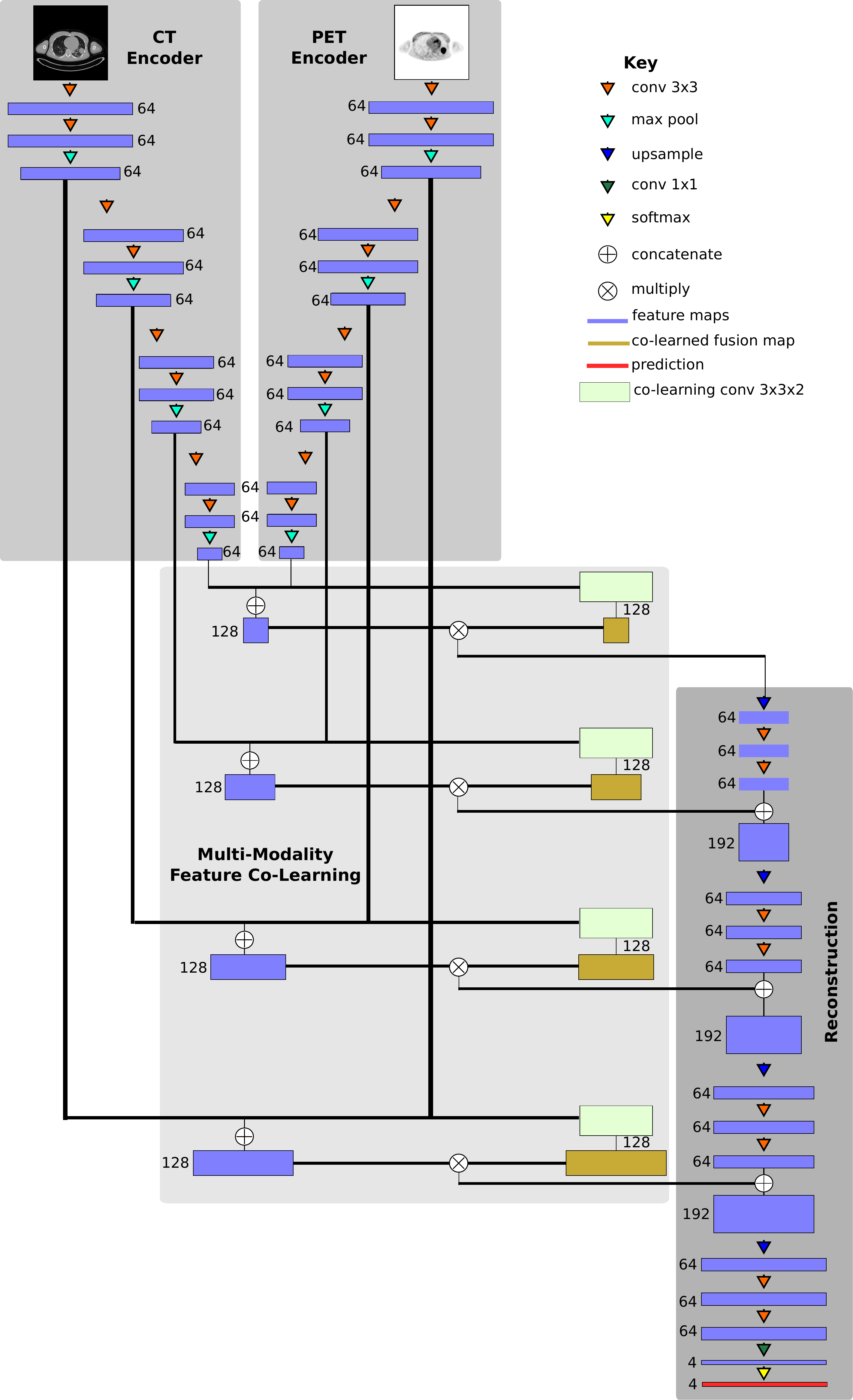}
    \caption{The architecture of our CNN, comprising two modality-specific encoders, a co-learning component, and a reconstruction component; the black lines indicate inputs to operations as skip connections between non-adjacent layers. The input to each modality-specific encoder is a 2D image slice of the corresponding modality.}
    \label{fig:arch}
\end{figure}

Fig.~\ref{fig:arch} shows the architecture of our proposed CNN; note that the number alongside each feature map in the figure refers to the number of output channels in the feature map. Our CNN comprises four main components: two encoders (one for each modality), one co-learning and fusion component, and a reconstruction component. The purpose of the two encoders is to derive the image features that are most relevant to each specific image modality; the input to each encoder is an axial 2D image slice. The co-learning component uses the modality-specific features produced by the encoders to derive a spatially varying fusion map to weight the modality-specific features at different locations. Finally, the reconstruction component integrates the modality-specific fused features across multiple scales to produce the final prediction map. The components are described in detail in the following subsections.

\subsection{Modality-Specific Encoders}
\label{sec:methods:encoder}
Our CNN contains an encoder for PET images and a separate encoder for CT images. The purpose of each encoder is to extract the visual features that are relevant to the input image modality. Thus, the encoders were designed with stacked convolutional layers in a similar manner to the deep CNNs that have achieved high accuracy in image classification tasks, e.g., AlexNet~\cite{krizhevsky2012_alexnet} and VGGNet~\cite{chatfield2014_devil,simonyan2014_vgg}. As shown in Fig.~\ref{fig:arch}, each encoder comprises four blocks that each contain two convolutional layers for feature map generation and a max pooling layer to down-sample the feature maps. 

A consequence of this stacked structure is that as the weights of each layer change, the distribution of the outputs they produce also change, potentially influencing the convolutional layers later in the network. During training, this means that even small changes in the weights of one layer may cascade and be amplified in deeper layers, requiring the layers to continuously adapt to new input distributions~\cite{shimodaira2000_inputdistshift}. As our CNN includes inputs from two different imaging modalities, the co-learning and reconstruction components will be affected by the cascading weight changes from both encoders, which will slow convergence and thus hinder the learning process.

Let $F = W \ast X + b$ be the output feature map of a convolutional layer where $X$ is the input to the convolution layer, $\ast$ is the convolution operation, $W$ is the learned weights of the convolution layer, and $b$ is the learned bias of the convolution layer. We use a batch normalization layer~\cite{ioffe2015_batchnorm} to normalize every dimension of the output feature map $F$ to a distribution with zero mean and unit variance, which acts to reduce the impact when the feature map is used as an input for subsequent convolutional layers.

We use the leaky rectified linear unit (Leaky ReLU) activation function~\cite{maas2013_leakyrelu} after feature map normalization:
\begin{equation}
	\psi_\alpha(\hat{f}) = 
    	\begin{cases} 
        	\hat{f}, & \hat{f} > 0 \\
        	\alpha\cdot\hat{f}, & \hat{f} \leq 0
    	\end{cases}
	\label{eq:leakyrelu}
\end{equation}
where $\hat{f}$ is a normalized feature and $\alpha$ is a parameter controlling the `leakiness' of the activation function, with the constraint that $0 < \alpha < 1$. The Leaky ReLU activation avoids the dead neuron problem that can occur with the standard ReLU function~\cite{maas2013_leakyrelu} where some weights in $W$ can be updated to a value where their training gradients are forever stuck at 0, thus preventing the weights from being updated in the future. The parameter $\alpha$ enables the introduction of a small non-zero gradient when $\hat{f} < 0$, thereby preventing the weights from being stuck at an unrecoverable value. For simplicity of notation, we refer to the output of a convolutional layer by $\hat{F} = \psi_\alpha(W \ast X + b)$ as the feature map generated from $X$ after convolution, batch normalization, and activation.

\subsection{Multi-modality Feature Co-Learning and Fusion}
\label{sec:methods:colearn}

\begin{figure}[!t]
	\centering
    \includegraphics[width=\columnwidth,keepaspectratio]{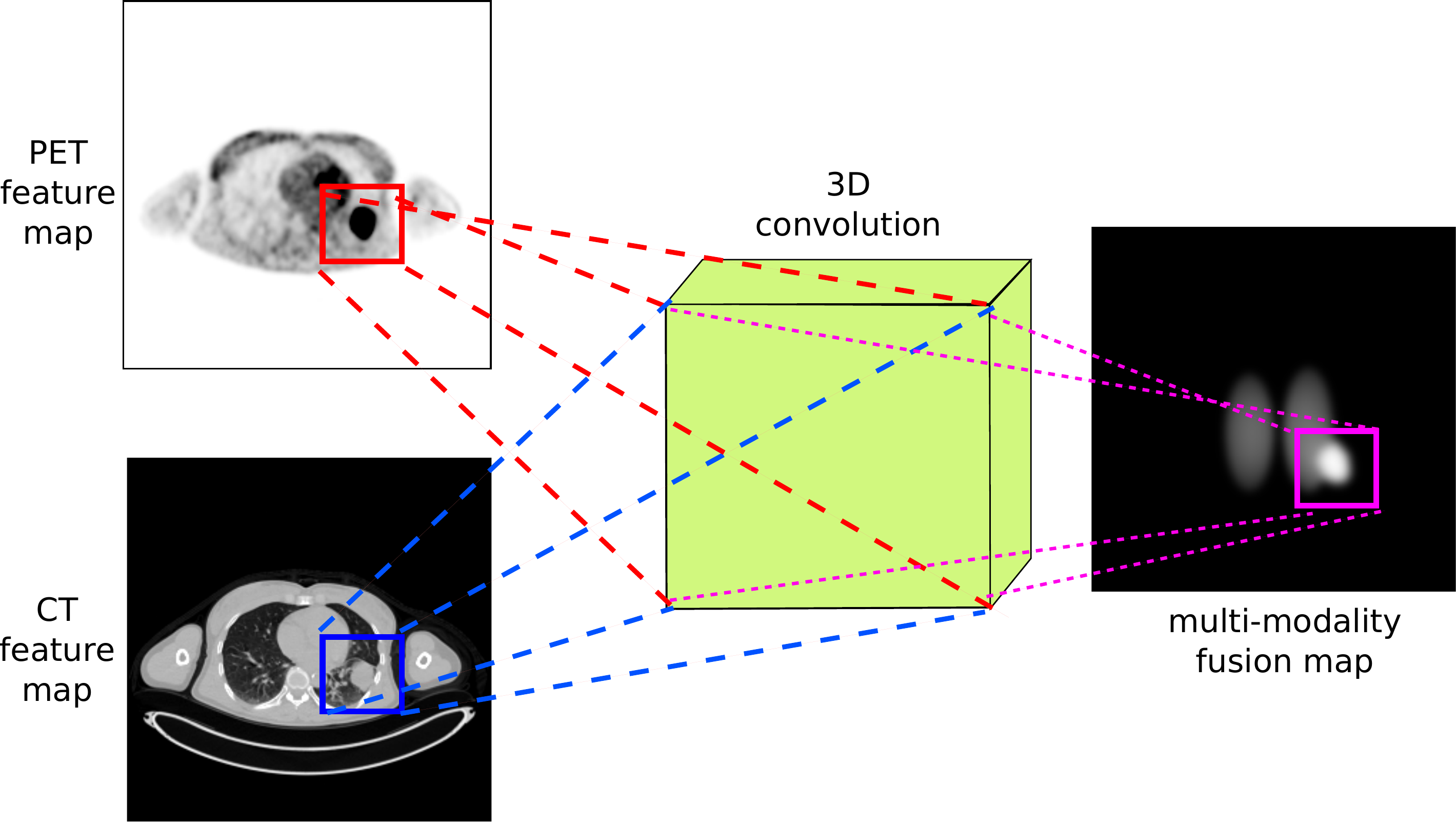}
    \caption{A conceptual example of the co-learning unit that learns to derive the fusion from the feature maps of each modality; for simplicity the figure shows co-learning on a single feature map channel.}
    \label{fig:colearn}
\end{figure}

The co-learning component consists of two parts: (i) a co-learning unit that is a CNN that learns to derive spatially varying fusion maps, and a (ii) fusion operation that uses the fusion maps to prioritize different features. Fig.~\ref{fig:colearn} shows a conceptual example of the function of the multi-modality co-learning unit. The inputs to the co-learning unit are two feature maps $\hat{F}_{CT}$ and $\hat{F}_{PET}$ (each from a block of one modality-specific encoder), each of size $w \times h \times c$ with $w$ width, $h$ height, and $c$ channels. These feature maps are stacked to form $\hat{X}_{multi}$, a $w \times h \times m \times c$ tensor with $m = 2$ number of modalities. The channels of $\hat{X}_{multi}$ are then convolved with the channels of a learnable 3D kernel $W_{multi}$ of size $k \times k \times m$, where $k$ is the width and height of the kernel, and $m = 2$ is the number of modalities. 

By performing a 3D convolution~\cite{ji2013_3dcnn} without padding the modality dimension, we obtain for a given channel $c$ a feature map with a singleton third dimension where the value at location $(x,y)$ is determined from the neighborhood of both $\hat{F}_{CT}(x,y)$ and $\hat{F}_{PET}(x,y)$:
\begin{equation}
\begin{split}
	(&W_{multi} \ast \hat{X}_{multi})(x,y) \\
    	& = \sum_{i}\sum_{j}\sum_{l} W_{multi}(i, j, l) \cdot \hat{X}_{multi}(x-i,y-j,l) 	
\end{split}
	\label{eq:conv-3d}
\end{equation}
We then squeeze the singleton third dimension to obtain an output feature map $W_{multi} \ast \hat{X}_{multi}$ of size $w \times h \times 2c$, the same width and height as the two modality-specific input feature maps $\hat{F}_{CT}(x,y)$ and $\hat{F}_{PET}(x,y)$ and double the number of channels, which is important for the weighting of modality-specific feature maps by the co-learned fusion maps as described below.

Our intention is that the co-learned fusion map controls the level of importance given to information from each modality at each location, in contrast to the global fusion ratio in PET-CT pixel intermixing~\cite{cai1999_dataintermix,quon2006_petctpilot,cheirsilp2015_petctthoracic}. Thus the co-learned fusion maps directly affect the input distribution of the learnable layers that immediately follow the co-learning unit. Hence, we do not normalize the output of the 3D convolution within the co-learning unit. As with the encoders (see Section~\ref{sec:methods:encoder}), we used a Leaky ReLU activation function to obtain the multi-modality co-learned fusion map:
\begin{equation}
	F_{fusion} = \psi\left(W_{multi} \ast \hat{X}_{multi} + b_{multi}\right)
    \label{eq:colearn}
\end{equation}
where $b_{multi}$ are the learned biases. Note that the multi-modality fusion map $F_{fusion}$ is obtained by the co-learning unit based on the spatial integration of the features from both modalities, since the 3D convolution operation considers the 3D neighborhood defined by the width, height, and modality of the stacked feature map $\hat{X}_{multi}$. 

\begin{figure}[!t]
	\centering
    \includegraphics[width=\columnwidth,keepaspectratio]{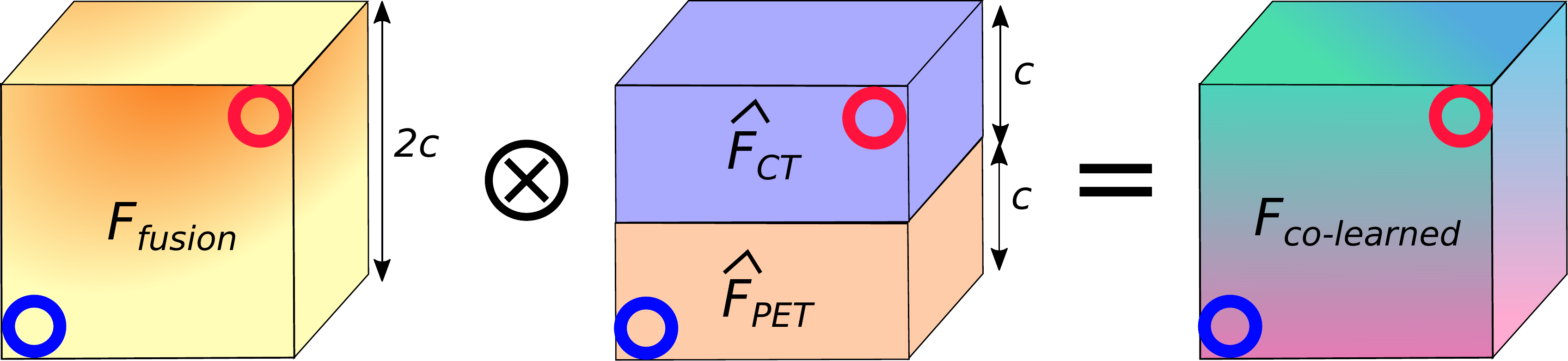}
    \caption{Multiplying the spatially varying fusion map ($F_{fusion}$) with the stacked modality-specific feature maps ($\hat{F}_{CT} \oplus \hat{F}_{PET}$) to generate a fused co-learned feature map ($F_{co-learned}$). Element-wise multiplication ensures that each  value in $F_{co-learned}$ is a weighted form of a  modality-specific feature. The red circles and blue circles indicate the element-wise multiplication of CT and PET features, respectively.}
    \label{fig:fusion}
\end{figure}

The fusion operation (depicted in Fig.~\ref{fig:fusion}) integrates the modality-specific feature maps according to the values (coefficients) in the multi-modality fusion map, as follows: 
\begin{equation}
	F_{fused} = F_{fusion} \otimes \left(\hat{F}_{CT} \oplus \hat{F}_{PET}\right)
\label{eq:fusion}
\end{equation}
where $F_{fused}$ is the fused co-learned feature map, $\oplus$ is the stacking operation, and $\otimes$ is an element-wise multiplication. This process merges the two modality-specific feature maps $\hat{F}_{CT}$ and $\hat{F}_{PET}$ and weights them by the co-learned multi-modality fusion map $F_{fusion}$, similar to pixel intermixing.  Our CNN (Fig.~\ref{fig:arch}) generates four fused feature maps, one for each pair of encoder blocks. These fused feature maps are passed to the reconstruction part of the CNN (see Section~\ref{sec:methods:reconstruct}).

\subsection{Reconstruction}
\label{sec:methods:reconstruct}

The reconstruction part of our CNN creates a prediction map of the ROIs within the PET-CT image. It does this by integrating the co-learned feature maps from different encoder blocks and upsampling them to the dimensions of the original inputs. Similar to the encoders, the reconstruction component comprises four blocks each with one upsampling layer and two convolutional layers.

The input to a reconstruction block is the output co-learned feature map from a co-learning unit stacked with the output of any prior reconstruction block. The upsampling layer first doubles the width and height of the stacked feature map using nearest neighbor interpolation to enable eventual reconstruction of the detected regions at the same scale as the original input; this formulation is similar to that of a deconvolutional layer~\cite{dumoulin2016_convmath} but does not require any striding operations. The following two convolutional layers merge and refine the information from the stacked modality-specific feature maps. The concept behind each reconstruction block is to generate higher dimensional feature maps that better correspond to the features for different ROIs by merging lower dimensional information with features that were fused from multiple image modalities. As with the modality-specific encoders (see Section~\ref{sec:methods:encoder}), we use batch normalization~\cite{ioffe2015_batchnorm} and Leaky ReLU~\cite{maas2013_leakyrelu} activations.

\begin{figure}[!t]
	\centering
    \includegraphics[width=\columnwidth,keepaspectratio]{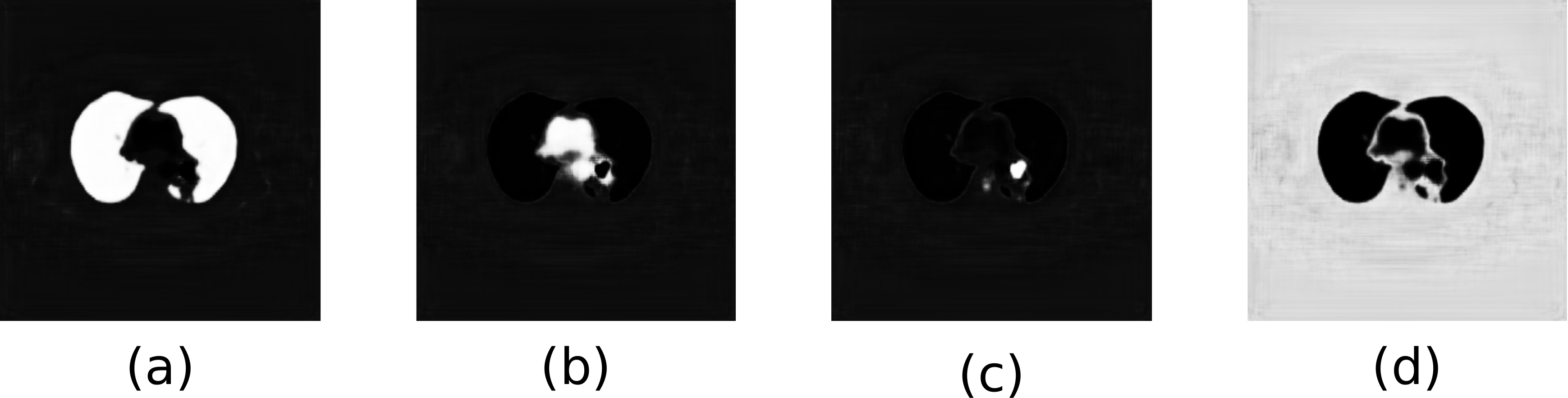}
    \caption{The probability maps generated by our CNN for the different regions; a higher (whiter) intensity implies a higher probability that a pixel belongs to a specific region. This example shows: (a) lung fields, (b) mediastinum, (c) tumors, and (d) `other' regions.}
    \label{fig:prediction}
\end{figure}

After the last reconstruction block, the output feature map has the same width and height as the input PET-CT image, with 64 channels in the third dimension. This is analogous to a final 64-dimensional feature vector for each pixel in the original image. We then use a 1$\times$1 convolution to map these feature vectors into $R+1$ feature maps, where $R$ is the number of ROIs. This obtains for each pixel a vector $o$ corresponding to the observed activations for each ROI class as well as an `other' class comprising all other image contents. We use the term `other' rather than `background' as this region encompasses the ‘true’ PET-CT background (areas of the image outside the field of view of the scanner that have a zero pixel value) as well as non-zero pixel areas within the image that are not of interest in the task (e.g., skin and subcutaneous fat of the chest wall, arms, etc.). Finally, we transform these observations into a probability or prediction map that corresponds to the likelihood of the pixel belonging to a particular class using the softmax function~\cite{fernandezcaballero2010_softmax}:
\begin{equation}
	p_i(o) = \left(\frac{e^{o_i}}{\sum_r^{R+1} e^{o_r}}\right)
	\label{eq:softmax}
\end{equation}
where $p_i(o)$ is the probability that the pixel with observation vector $o$ belongs to the region $i$, $o_i$ is the $i$-th element of vector $o$ and is the activation corresponding to region $i$. The `other' class (and hence the summation for $R+1$ regions in the denominator of Equation~\ref{eq:softmax}) is necessary to formulate the final output of our co-learning CNN as a set of probability maps. The `other' class probability map ensures the sum of probabilities for each pixel has a total of 1 by capturing the probability that a pixel does not belong to any of the ROIs. The use of an additional class to compute the probability of non-ROI regions is a standard formulation that has been used in prior CNN research~\cite{shelhamer2017_fcn}. Fig.~\ref{fig:prediction} is an example of the probability maps generated for the $R=3$ classes used in our experiments (lung fields, mediastinum, tumors), and the `other' class.

\subsection{Network Training}
\label{sec:methods:train}

We trained our CNN using stochastic mini-batch stochastic gradient descent with momentum~\cite{sutskever2013_sgdmom} using the following loss function and training parameters. To improve the robustness of our training and to avoid overfitting we applied data augmentation through the standard technique of random cropping and flipping of training samples~\cite{chatfield2014_devil,kumar2017_jbhi}.

\subsubsection{Loss Function}
\label{sec:methods:train:loss}
We modified the well-established categorical cross-entropy loss function for training our CNN. Let $o \in O$ be the set of pixel observations in an image $O$ and $T(o)$ be the true class of $o$, from a set of $R+1$ ROIs. Then our loss is given by:
\begin{equation}
	L(O) = \frac{1}{|O|}\sum_o S(o)E(o) + \lambda\sum_j (w_j)^2
    \label{eq:loss}
\end{equation}
where
\begin{equation}
	S(o) = 1-\frac{n_{T(o)}}{\sum_r^{R+1} n_r}
    \label{eq:loss-scale}
\end{equation}
is the class specific scaling, and
\begin{equation}
	E(o) = -\sum_i^{R+1} y_i(o)\log \left(p_i(o)\right)
    \label{eq:cross-entropy}
\end{equation}
is the cross-entropy loss~\cite{anthimopoulos2016_cnnild}. Under this formulation, $n_{T(o)}$ is the number of pixels in the true class of $o$, $n_r$ is the number of pixels in class $r$, $y_i(o)$ is an indicator function that is 1 when $i = T(o)$ and 0 otherwise, $p_i(o)$ is defined by Equation~\ref{eq:softmax}, $\lambda$ is the regularization strength, and $w_j$ is the $j$-th weight in $\mathcal{W}$, the set of all weights in the CNN. The distribution of the number of pixels in each class varies depending on the particular ROI (e.g., there are many more lung pixels than there are tumor pixels). As such, $S(o)$ in Equation~\ref{eq:loss} acts as a scaling coefficient for the cross entropy loss $E(o)$; this formulation is designed to reduce any bias that may be caused by ROIs with different sizes (e.g., tumor ROIs are often much smaller than lung fields)~\cite{ronneberger2015_unet}. The final term in Equation~\ref{eq:loss} is a regularization to reduce overfitting. Our aim was to ensure that the convolution kernel weights (and as a consequence, the features) corresponding to one modality did not overpower the weights (and the features) of the other. As such, we used an $L_2$-regularization, which acts to prioritize lower weights across the entirety of $\mathcal{W}$~\cite{han2015_l2neuralnet}.

\subsubsection{Parameter Selection}
\label{sec:methods:train:params}

We empirically derived the parameters using a two-fold cross-validation approach on the training data (see Section~\ref{sec:methods:materials}). Table~\ref{tab:params} lists the parameters used for our training. Further information on our parameter validation is provided in Section~SI of the Supplementary Materials.

\begin{table} [!t]
	\renewcommand{\arraystretch}{1.3}
	\centering
    \caption{CNN Architecture and Training Parameters}
    \label{tab:params}
    \begin{tabular} {| r | c |}
    	\hline
    	{\bfseries Architecture Parameter} & {\bfseries Value} \\
        \hline
        2D convolution kernel size & 3$\times$3 \\        
        convolution stride & 1 \\
        max pool size & 2$\times$2 \\
        pool stride & 2 \\
        number of channels ($c$) & 64 \\
        3D convolution kernel size & 3$\times$3$\times$2 \\
        \hline
        \hline
    	{\bfseries Training Parameter} & {\bfseries Value} \\
        \hline
        ReLU leakiness ($\alpha$) & 0.1 \\
        regularization strength ($\lambda$) & 0.1 \\
        learning rate & 0.0001 \\
        momentum & 0.9 \\
        batch size & 5 \\
        \# epochs & 500 \\
        \hline
    \end{tabular}
\end{table}

\subsection{Experimental Design}
\label{sec:methods:eval}

We implemented our CNN using Tensorflow 1.4~\cite{abadi2016_tensorflow} on a machine running Ubuntu 14.04 with CUDA 8.0 and CuDNN~\cite{chetlur2014_cudnn}. Training was performed on an 11GB NVIDIA GTX 1080 Ti. A link to our code can be found in Section~SV of the Supplementary Materials.

For robustness and to reduce bias, we used a 5-fold cross validation evaluation approach. For each fold, we used the same training and test datasets for our method and all baselines. We used greyscale inputs for both modalities, as was common in the baseline fusion strategies ~\cite{zhang2015_neuromm,bradshaw2018_vgg3chanpetct,teramoto2016_petctpulmnod} and other multi-modality CNN research~\cite{bi2017_petctclass,liu2018_mrctattencorr,zhang2015_neuromm,tseng2017_crossmodconv}. For all experimental comparisons with baseline methods (see below), we computed the $p$-value with the two-sample $t$-test.

We used two main evaluation tasks to demonstrate the usefulness of our method: the detection and segmentation of ROIs in PET-CT images of lung cancer. We used lung cancer as the target disease because the fusion requirements for tumors within the lung field are well-established: the lungs require mainly CT and the tumors require mostly PET with some CT within the lung area. This well-known requirement could be used to validate whether our co-learning CNN could derive the appropriate spatially varying fusion maps for tumors within the lung. This would indicate that our CNN could learn to produce spatially varying fusion maps for disease in other anatomical locations. As such, our evaluation scenarios required the detection and segmentation of disease within the lung field as well as in the mediastinum and hilar nodes. The specific experiments are detailed below.

\subsubsection{Comparison with Fusion Baselines (Region Detection and Segmentation)}
\label{sec:methods:eval:detection}
We compared our fusion method to several fusion baseline strategies. To limit the number of variable changes in our experimentation, for all fusion baselines we used a similar architecture as in our method (Fig.~\ref{fig:arch}), replacing the co-learning component with a fusion strategy from the literature. The baselines were:
\begin{itemize}
	\item A multi-branch (MB) CNN, implementing a fusion strategy where each modality was processed separately and the outputs from each modality were combined~\cite{teramoto2016_petctpulmnod,vantulder2017_sharedcrossmod,zhong2018_petctunet}. The CNN was similar to the architecture in Fig.~\ref{fig:arch} with no co-learning component. 
    \item A multi-channel (MC) input CNN, implementing a fusion strategy where each modality was treated as different channels of a single input~\cite{zhang2015_neuromm,bradshaw2018_vgg3chanpetct}. The CNN was similar to a single encoder form of the architecture in Fig.~\ref{fig:arch}, with no co-learning component and the CT and PET modalities input as separate channels.
    \item A fused (FS) input CNN, implementing a strategy where the input was a PET-CT image that had already been fused via pixel-intermixing~\cite{xu2018_petctbone}. The CNN was similar to a single encoder form of the architecture in Fig.~\ref{fig:arch} with no co-learning component.
\end{itemize}
We measured our CNN's effectiveness in detecting the foreground ROIs (lungs, mediastinum, tumors) and the `other' image contents. This was used to verify if parts of the `other' region could be misinterpreted as an ROI or vice versa (e.g., an area of the image with high intensity PET noise being mistaken as a tumor). Our comparisons used the following metrics calculated from the overlap of the detected region with the ground truth (GT): precision, sensitivity (recall), specificity, and accuracy. We also measured the segmentation quality of the predicted foreground regions using the Dice score.

\subsubsection{Comparison with Lung Tumor Segmentation Baseline}
\label{sec:methods:eval:segmentation}
We compared our co-learning CNN with two recent deep learning methods for PET-CT lung tumor segmentation:
\begin{itemize}
    \item A tumor co-segmentation~\cite{zhong2018_petctunet} method. We used the publicly available source code. We trained the baseline CNN on our dataset using the baseline's default parameters except for the batch size and the number of training epochs, which we increased to match the training scheme of our co-learning CNN (see Table~\ref{tab:params}). The baseline segmented the CT and PET images separately, which could be merged for a PET-CT segmentation; we have divided our results accordingly.
    \item A variational tumor segmentation method~\cite{li2019_dlpetctseg}. The source code was not publicly available and thus we implemented the baseline according to the details provided in the paper. This baseline used a U-Net to coarsely identify tumor ROIs on CT, which were then refined using the PET data and a fuzzy variational model.
\end{itemize} The baselines were only designed to segment the tumor and hence we only compared the Dice score for the tumor ROI. Both baselines were designed for inputs that were image patches centered around the tumor region. To ensure fair comparisons, we performed separate experiments using patch inputs as well as full PET-CT slices ($256\times256$ pixels).

\subsubsection{Evaluation of the Fusion Effects}
\label{sec:methods:eval:fusion}
We extracted the feature fusion maps produced by the co-learning unit (see Section~\ref{sec:methods:colearn}) to examine the fusion ratios that were produced for an image with a tumor inside the lung field. This experiment was undertaken to confirm whether the fusion ratios that were automatically derived matched the well-established expectations. Further analysis on a heterogeneous tumor was performed for the Supplementary Materials (see Section~S3). We also visually compared the results derived using our co-learning CNN's spatially varying fusion to the results derived from using different uniform fusion ratios with the FS architecture. We used uniform fusion ratios that included mainly anatomical information (30\% PET with 70\% CT),  equal information (50\% PET and CT), and mainly functional information (70\% PET with 30\% CT).

\begin{table*} [!t]
	\renewcommand{\arraystretch}{1.3}
	\centering
	\begin{threeparttable}	
    \caption{Comparison of fusion CNNs on detection and segmentation of different ROIs}
    \label{tab:res_compare}
    \begin{tabular} {| c || c || c | c | c | c || c |}
    	\hline
		\multicolumn{2}{| c ||}{\bfseries Metrics}	& \multicolumn{4}{ c ||}{\bfseries Detection [Mean $\pm$ Standard Deviation \%] } & \multicolumn{1}{ c |}{\bfseries Segmentation [Mean] }	\\
        \hline
        \hline
        \multicolumn{1}{| c ||}{\bfseries ROI} & {\bfseries CNN} & Precision & Sensitivity & Specificity & Accuracy & Dice (\%)\\
        \hline
        \hline
        \multirow{4}{*}{\rotatebox[origin=c]{90}{lungs}}
        	& MB &80.95 $\pm$ 6.35* &99.38 $\pm$ \mbox{ }{\bfseries0.88}* &98.55 $\pm$ 0.50* &98.60 $\pm$ 0.46* &89.09*\\
            & MC &78.84 $\pm$ 9.68* &99.40 $\pm$ \mbox{ }1.02\mbox{ } &98.11 $\pm$ 3.24* &98.19 $\pm$ 3.08* &87.51*\\
            & FS &79.11 $\pm$ 8.64* &99.06 $\pm$ \mbox{ }1.23* &98.38 $\pm$ 0.66* &98.42 $\pm$ 0.62* &87.68*\\
            & our method &{\bfseries85.31 $\pm$ 6.07}\mbox{ } &{\bfseries99.49} $\pm$ \mbox{ }0.93\mbox{ } &{\bfseries98.95 $\pm$ 0.43}\mbox{ } &{\bfseries98.98 $\pm$ 0.39}\mbox{ } &{\bfseries91.73}\mbox{ }\\
        \hline
        \multirow{4}{*}{\rotatebox[origin=c]{90}{mediastinum}}
        	& MB &59.75 $\pm$ 15.16* &93.28 $\pm$ \mbox{ }9.21\mbox{ } &98.91 $\pm$ 0.60* &98.83 $\pm$ 0.58* &71.57*\\
            & MC &58.15 $\pm$ 15.45*  &{\bfseries94.44 $\pm$ \mbox{ }7.52}\mbox{ } &98.85 $\pm$ 0.58* &98.78 $\pm$ 0.57* &70.60*\\
            & FS &58.58 $\pm$ {\bfseries14.79}* &90.67 $\pm$ 14.67* &98.89 $\pm$ 0.58* &98.78 $\pm$ 0.59* &70.16*\\
            & our method &{\bfseries64.54} $\pm$ 16.14\mbox{ } &94.15 $\pm$ \mbox{ }8.81\mbox{ } &{\bfseries99.12 $\pm$ 0.51}\mbox{ } &{\bfseries99.04 $\pm$ 0.51}\mbox{ } &{\bfseries 75.25}\mbox{ }\\
        \hline
        \multirow{4}{*}{\rotatebox[origin=c]{90}{tumors}}
        	& MB &56.57 $\pm$ 29.05* &68.43 $\pm$ 33.81* &99.86 $\pm$ {\bfseries0.13}* &99.79 $\pm$ 0.18* &52.16*\\
            & MC &56.83 $\pm$ 29.63* &66.65 $\pm$ 35.51* &99.86 $\pm$ 0.14* &99.80 $\pm$ 0.16* &49.31*\\
            & FS &54.26 $\pm$ {\bfseries28.27}* &72.21 $\pm$ 32.76* &99.84 $\pm$ 0.17* &99.79 $\pm$ 0.17* &53.07*\\
            & our method &{\bfseries64.56} $\pm$ 29.61\mbox{ } &{\bfseries79.97 $\pm$ 28.26}\mbox{ } &{\bfseries99.89 $\pm$ 0.13}\mbox{ } &{\bfseries99.85 $\pm$ 0.14}\mbox{ } & {\bfseries 63.85}\mbox{ }\\
        \hline
        \hline
        \multirow{4}{*}{\rotatebox[origin=c]{90}{foreground}}
        	& MB &75.06 $\pm$ {\bfseries6.78}* &97.43 $\pm$ 2.58* &99.12 $\pm$ 0.29* &99.08 $\pm$ 0.31* & 84.64*\\
            & MC &73.04 $\pm$ 9.29* &97.72 $\pm$ {\bfseries1.90}* &98.96 $\pm$ 1.07* &98.92 $\pm$ 1.05* & 83.19*\\
            & FS &73.60 $\pm$ 7.69* &97.03 $\pm$ 2.61* &99.05 $\pm$ 0.34* &99.00 $\pm$ 0.34* & 83.48*\\
            & our method &{\bfseries79.66} $\pm$ 7.53\mbox{ }  &{\bfseries97.94} $\pm$ 1.93 &{\bfseries99.33 $\pm$ 0.24}\mbox{ } &{\bfseries99.29 $\pm$ 0.25}\mbox{ } & {\bfseries87.66}\mbox{ }\\
        \hline
        \hline \multirow{4}{*}{\rotatebox[origin=c]{90}{`other'}}
        	& MB &99.90 $\pm$ 0.13\mbox{ } &97.32 $\pm$ 0.93* &98.83 $\pm$ 1.96\mbox{ } &97.45 $\pm$ 0.84* &---\\
            & MC &{\bfseries99.91} $\pm$ 0.14* &96.80 $\pm$ 3.33* &{\bfseries99.05 $\pm$ 1.30}* &96.98 $\pm$ 3.13* &---\\
            & FS &99.87 $\pm$ 0.19* &97.11 $\pm$ 1.06\mbox{ } &98.55 $\pm$ 2.29* &97.24 $\pm$ 0.99* &---\\
            & our method &99.90 $\pm$ {\bfseries0.11}\mbox{ } &{\bfseries97.94 $\pm$ 0.76}\mbox{ } &98.82 $\pm$ 1.33\mbox{ } &{\bfseries98.02 $\pm$ 0.71}\mbox{ } &---\\
        \hline
    \end{tabular}
    \begin{tablenotes}
    	\item[*] $p < 0.05$, in comparison to our method as derived from a $t$-test.
    	\item MB: multi-branch CNN, MC: multi-channel CNN, FS: fused input CNN
    \end{tablenotes}
    \end{threeparttable}
\end{table*}

\section{Results}
\label{sec:res}

Table~\ref{tab:res_compare} shows the comparison of our co-learning CNN with the baseline fusion methods on ROI detection and segmentation experiments. The data are presented individually for each of the three ROI, for all foreground ROI collectively, and separately for the non-ROI `other' region. In the detection experiments, our co-learning method has higher mean accuracy when compared to all baselines for all individual ROIs and for the foreground. The improvement in accuracy offered by our method is statistically significant ($p < 0.05$) for all ROIs. In ROI and foreground detection, our co-learning fusion method improves upon all baselines in 15 of the 16 metrics and 14 of these improvements are statistically significant compared to all baselines (all 15 improvements are statistically significant over at least one baseline). The largest overall improvement was in the precision metric, indicating that our method resulted in an increase in the ratio of true positives to false positives. In the segmentation experiments, our co-learning CNN had a significantly higher Dice score ($p < 0.05$) than all baseline fusion CNNs for all foreground ROIs individually and collectively.  

\begin{figure*}[!t]
	\centering
    \includegraphics[width=0.8\textwidth,keepaspectratio]{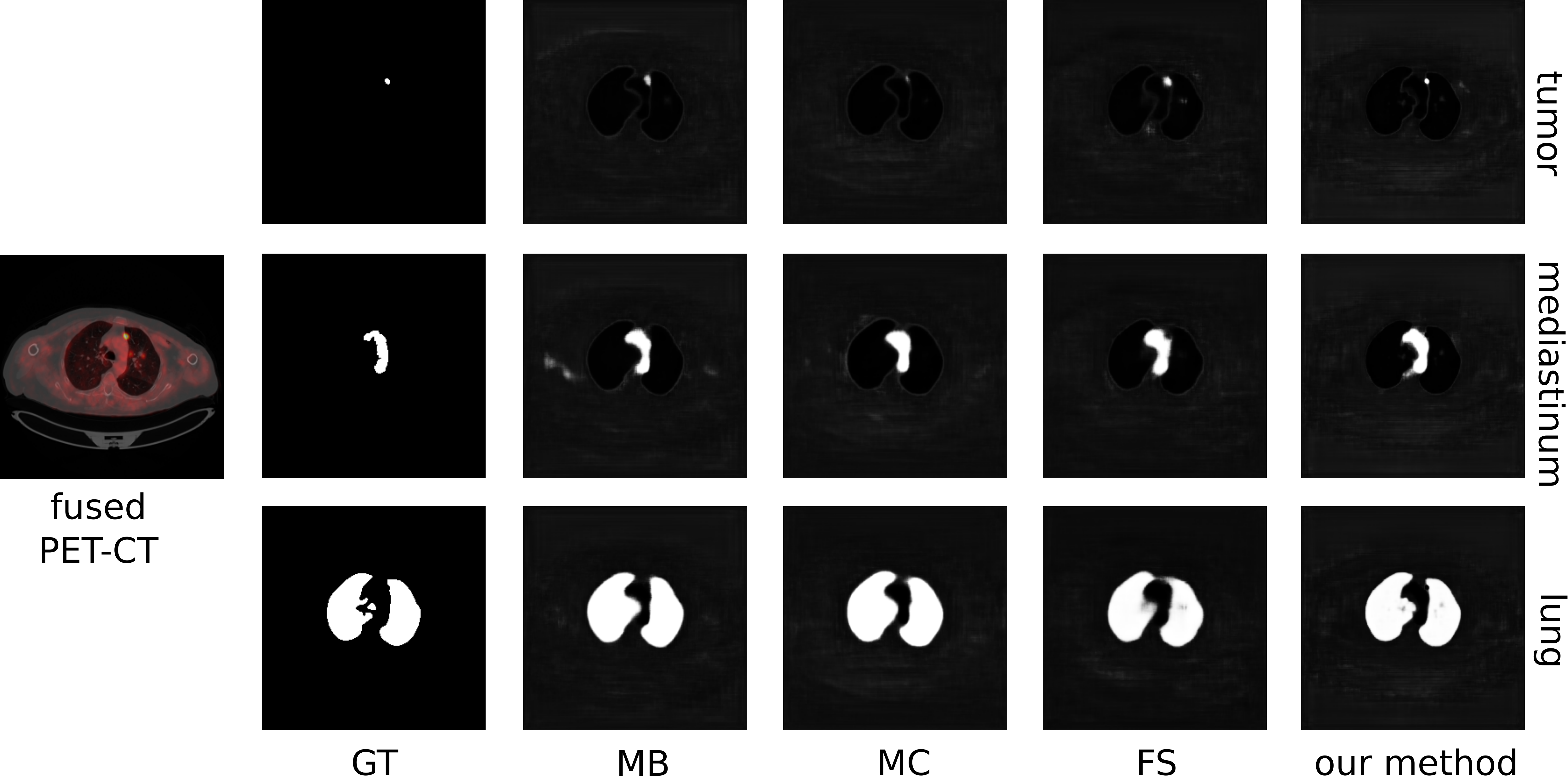}
    \caption{Visual comparison of the results obtained by our method compared to the baselines and the ground truth (GT). For clarity, we show on the left the fused form of the original PET-CT image, with a color lookup table applied to the PET modality.}
    \label{fig:res_compare_mm}
\end{figure*}

\begin{figure*}[!t]
	\centering
    \includegraphics[width=\textwidth,keepaspectratio]{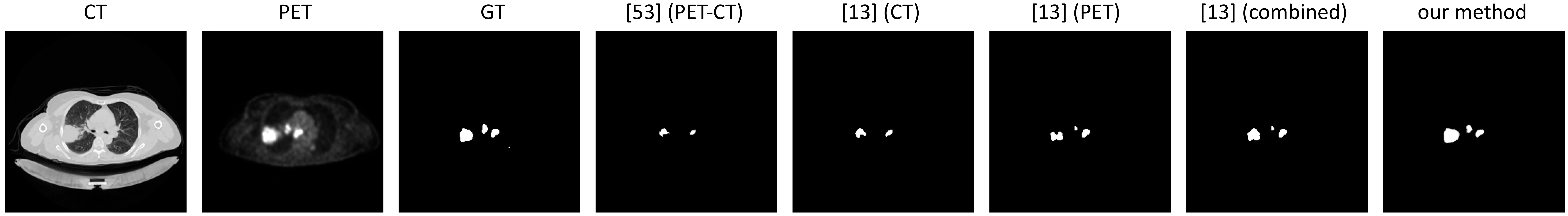}
    \caption{Visual comparison of the segmentation results obtained by our method compared to the baseline tumor segmentation methods for an PET-CT image slice with tumors in the lung and the mediastinum.}
    \label{fig:res_seg}
\end{figure*}

\begin{figure}[!h]
	\centering
    \includegraphics[width=0.9\columnwidth,keepaspectratio]{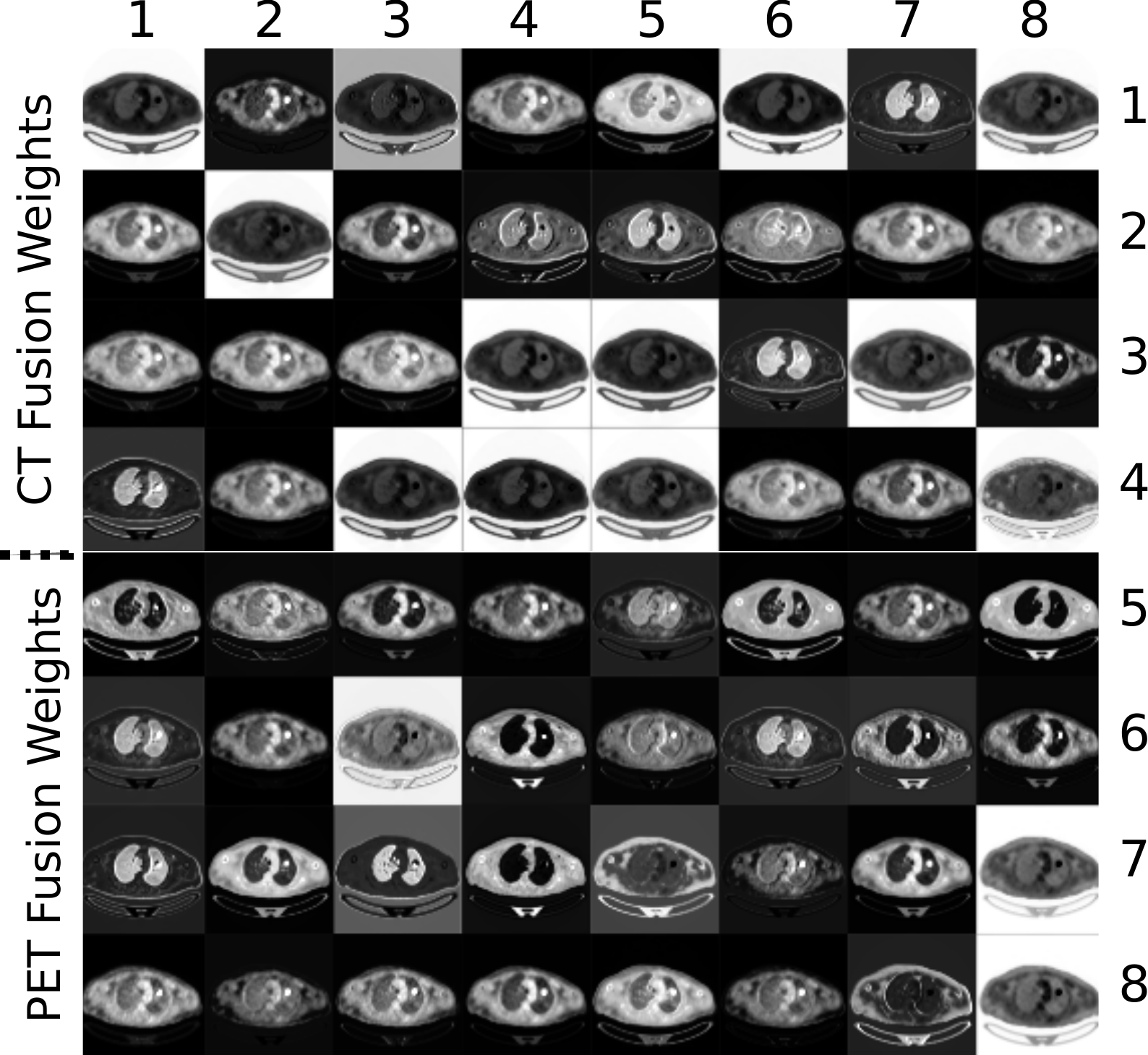}
    \caption{Fusion maps obtained by the first co-learning unit in our CNN. For better visualization, each map has been independently normalized. Areas with higher intensity values represent fusion weights that are relatively more important than areas with lower intensity values.}
    \label{fig:res_fusion}
\end{figure}

\begin{table} [!t]
	\renewcommand{\arraystretch}{1.3}
	\centering
	\begin{threeparttable}	
    \caption{Comparison with Tumor Segmentation Baseline}
    \label{tab:res_dice_compare}
    \begin{tabular} {| c | r || c |}
        \hline
         {\bfseries Method} &  \multicolumn{1}{ c ||}{\bfseries Input} & {\bfseries Mean Dice (\%) } \\
        \hline
        \multirow{2}{*}{Li et al.~\cite{li2019_dlpetctseg}}
            & PET-CT patch & \mbox{ }36.45 *\\
            & PET-CT slice & \mbox{ }\mbox{ }7.61 *\\ 
        \hline
        \multirow{6}{*}{Zhong et al.~\cite{zhong2018_petctunet}}
            & CT patch & \mbox{ }45.62 *\\
            & PET patch & \mbox{ }62.37 *\\
            & combined patch & \mbox{ }63.09 *\\
        	& CT slice &  \mbox{ }12.09 *\\
            & PET slice &  \mbox{ }62.29 *\\
            & combined slice &  \mbox{ }60.13 *\\
            \hline
            our method & PET-CT slice &  {\bfseries63.85}\mbox{ }\mbox{ }\\
        \hline
    \end{tabular}
    \begin{tablenotes}
        \item[*] $p < 0.05$, in comparison to our method.
    \end{tablenotes}
    \end{threeparttable}
\end{table}

\begin{figure}[!t]
	\centering
    \includegraphics[width=0.8\columnwidth,keepaspectratio]{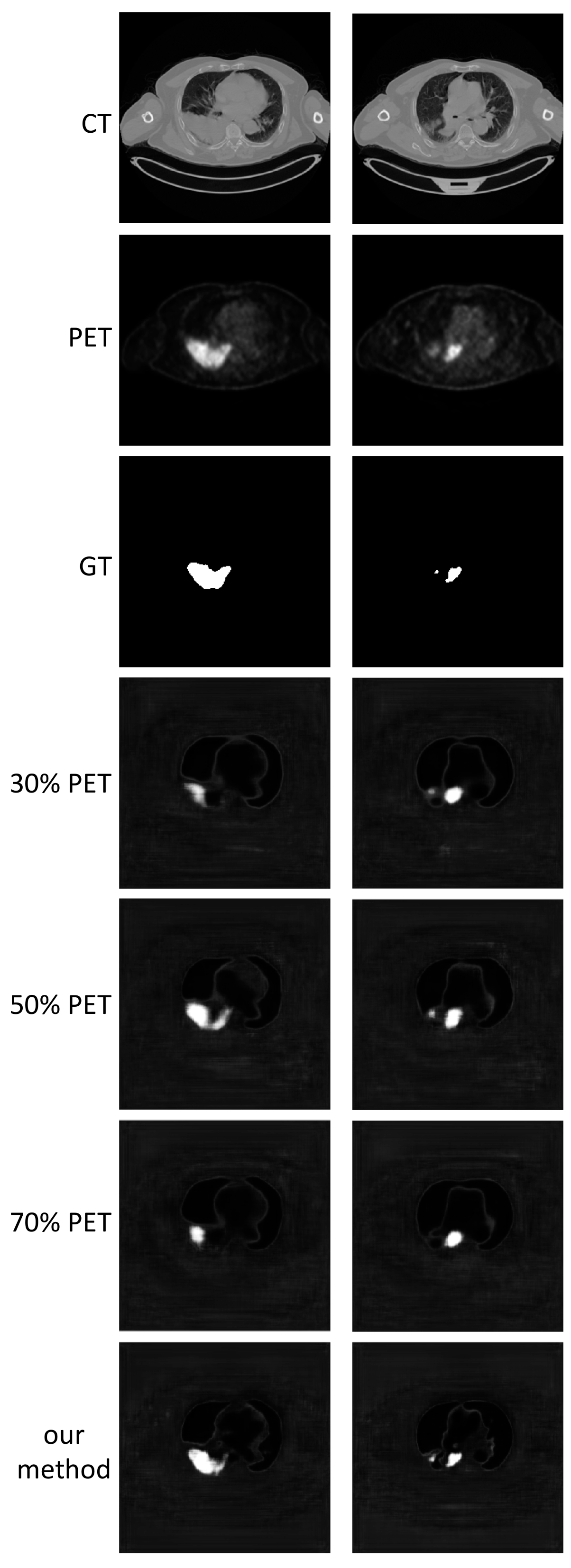}
    \caption{Visual comparison of the results obtained with co-learned fusion and uniform fusion using three different fusion ratios.}
    \label{fig:res_fusion_compare}
\end{figure}

Fig.~\ref{fig:res_compare_mm} is a visual comparison of the ROIs detected by our method and by the baselines; a larger version is included as Fig.~S3 in the Supplementary Materials. The figure shows that our method consistently detected regions that were a similar size to the ground truth. In contrast, the MC baseline detected fewer pixels (as shown by the tumor region) while the MB and FS baselines detected more pixels than within the region. In particular, the MB CNN gave pixels within the chest wall a high probability of being within the mediastinum.

Table~\ref{tab:res_dice_compare} is a comparison of the tumor segmentation performed by our co-learning CNN with two recently published PET-CT lung tumor segmentation techniques. The tumors segmented by our co-learning CNN have a significantly ($p < 0.05$) higher Dice score that the baselines. 

Fig.~\ref{fig:res_seg} is a visual comparison of the segmentation results for a PET-CT image slice with tumors within the lung field as well as in the mediastinum; a larger version is included as Fig.~S4 in the Supplementary Materials. The figure shows that our method was able to segment three tumors in different locations across the slice, although they are all slightly over-segmented relative to the GT. In contrast, baseline methods had a tendency to under-segment the tumors.

Fig.~\ref{fig:res_fusion} depicts the co-learned fusion maps that were derived for an image with a single tumor; a larger version is included as Fig.~S5 in the Supplementary Materials. In the figure, each feature map channel has been independently normalized so that their real valued pixels could be viewed in the paper. In any particular channel, a higher absolute intensity implies a greater importance placed on that pixel during fusion. The figure shows how different information is prioritized differently for each region. For example, the 7th CT fusion channel (row 1, column 7) places a greater emphasis on the lungs while the 26th PET fusion channel (row 8, column 2) places the greatest emphasis on the tumor. The figure also indicates that the fusion weights are derived from features of both modalities. For example, the 7th CT fusion channel (row 1, column 7) emphasizes the lungs \emph{including} the area that contains the tumor. Meanwhile, the 13th CT fusion channel (row 2, column 5) also emphasizes the lungs but \emph{de-emphasizes} the area containing the tumor. Further analyses are included in Section~SIII of the Supplementary Materials.

Fig.~\ref{fig:res_fusion_compare} is a visual comparison of the results obtained by our co-learning CNN to the results obtained with uniform fusion; a larger version is included as Fig.~S6 in the Supplementary Materials. The figure shows that our CNN  has visually consistent tumor detection across both studies. In contrast, the figure visually shows that a uniform fusion ratio may not be optimal for different studies; equal (50\%) PET and CT is the better ratio for detecting the tumor in the study in the left column, while reduced (30\%) PET is the better ratio for the detecting the tumors in study in the right column. The uniform fusion results were sensitive to the fusion ratio and the specific PET-CT images being processed, and across different studies produced probability maps that either missed tumors or overestimated the tumor area.

\section{Discussion}
\label{sec:discuss}

Our main findings are that our co-learning method improved foreground detection accuracy, provided a more consistent detection of regions when compared with the baseline fusion CNNs, and performed better than baseline tumor segmentation methods. We attribute these findings to the ability to derive a spatially varying fusion map that more precisely integrates functional and anatomical visual features across different locations in PET-CT images. 

\subsection{Comparison with Baseline Fusion CNNs}
\label{sec:discuss:comparisonfusion}

Our co-learning CNN achieved a higher detection precision, sensitivity, specificity, and accuracy than the MB CNN for fusion across all foreground ROIs individually and collectively. Our explanation for this outcome is that the design of our CNN explicitly fuses features at multiple scales through the multiple co-learning units, which prevents information loss that can occur from the standard pooling (downsampling) operations used for feature map dimensionality reduction in CNNs. In contrast, the MB CNN implements a late fusion approach in which modality-specific feature maps are merged just prior to the reconstruction, meaning that useful complementary information could possibly have already been lost. An examination of Fig.~\ref{fig:res_compare_mm} shows that the MB CNN tends to have larger predicted regions compared to the GT (e.g. larger tumor area, extra regions in mediastinum), indicating that the lost complementary information makes the MB CNN less precise.

The MC CNN implements an early fusion approach in which no modality-specific feature maps are derived and where the first convolutional layer combines both modalities to derive fused feature maps. However, as indicated by the metrics in Table~\ref{tab:res_compare} and the images in Fig.~\ref{fig:res_compare_mm} this tends to prioritize information from some modalities at the expense of information from the other modality. The clearest example is in the less precise detection of the tumor region, which is barely noticeable in Fig.~\ref{fig:res_compare_mm}; only the part of the tumor with peak SUV (highest radiotracer uptake) is detected and the less subtle tumor regions are missed altogether.

The FS baseline is another variant of early fusion; the PET and CT modalities are pre-fused via pixel intermixing and the intermixed image is used as the input. It shares a similar weakness to the MC CNN in that the pre-fusion acts to prioritize information from one modality at the expense of the others, resulting in good precision for the lungs (79.11\% in Table~\ref{tab:res_compare}) but much lower precision for non-lung ROIs. Examination of Fig.~\ref{fig:res_compare_mm} shows that the tumor and mediastinum regions detected by the FS CNN are larger than the GT, indicating that there are a greater number of false positives. Fig.~\ref{fig:res_fusion_compare} shows that our co-learning CNN produces more consistent ROI detections than the FS baseline, which is sensitive to the selection of the uniform fusion ratio parameter. This suggests that the spatially varying fusion derived within our co-learning CNN may be more robust than a pre-determined fusion parameter setting.

All fusion baselines and our co-learning CNN had consistently high specificity for the foreground ROIs, individually and collectively (Table~\ref{tab:res_compare}). This is expected due to the large `other' region in the images, which would provide more pixels to learn the characteristics of true negative samples. This analysis is supported by the high precision ($> 99.8\%$) achieved by all methods in detecting the `other' region, correctly recognizing that `other' regions are distinct from the 3 foreground ROIs. The effectiveness of the CNNs in separating the `other' region from the ROIs indicates that there is a low likelihood that portions of the `other' region could be misinterpreted as one of the ROIs or vice versa (e.g., an area outside the lung with high intensity PET noise being detected as a tumor).

Our co-learning CNN predicts ROIs with a significantly higher ($p < 0.05$) Dice score  (Table~\ref{tab:res_compare}) when compared to the baseline fusion methods in an image segmentation task. We note that the MC baseline is similar to the fully convolutional network~\cite{shelhamer2017_fcn} (MC has multiple upsampling stages rather than one), and hence the results indicate that our approach may be comparable to existing CNNs for image segmentation. CNN-based medical image segmentation techniques often include postprocessing steps to refine the coarse or soft outputs produced by the CNNs~\cite{kamnitsas2017_segcnncrf,christ2016_liverlesionfcncrf}; applying similar postprocessing steps would enhance the segmentation quality of our co-learning CNN. We suggest that the coarse outputs from our co-learning CNN may be directly used as the initial inputs for structure-specific analysis. Alternatively, our CNN could be extended and retrained for end-to-end use in specific computer-aided diagnosis applications.

We note that the Dice score achieved by our method for the lung (approximately 91\%) is generally lower than the state-of-the-art results achieved in lung segmentation competitions (approximately 95\%)~\cite{yang2018_lungautosegmentation}. One reason for this is that our co-learning CNN design and training loss function was focused on general fusion rather than a fusion that was optimized for segmentation, resulting in coarse boundaries that did not match the exact ROI. However, the high Dice score for the lung achieved by our method indicates that it is still able to capture most of the lung area. We suggest that the results could be further improved by training our CNN with a segmentation-specific loss (such as a scaled multi-class segmentation Dice loss) and performing multi-scale refinement (such as with RefineNet~\cite{lin2017_refinenet}).

\subsection{Tumor Segmentation Performance}
\label{sec:discuss:comparisonsegmentation}

Our method has a significantly higher Dice score  ($p < 0.05$) than the baseline tumor segmentation methods~\cite{zhong2018_petctunet,li2019_dlpetctseg}. The baseline methods produced lower Dice scores when using slices as inputs compared to using patches as inputs. The reason for this outcome is that like other recent PET-CT lung tumor segmentation techniques~\cite{zhao2018_tumorpetctseg,zhong2019_simultaneouscosegpetct}, the baselines were designed for inputs where the tumors were cropped from the full image and centered within an image patch. The results indicate that it is likely that the baselines were unable to identify the tumors when the image contained more varied anatomical and functional information from the full image slice, such as with tumors outside the lung fields (e.g., poor segmentation of tumors in the mediastinum or hilar nodes in Fig.\ref{fig:res_seg}). Our CNN had a comparable Dice score to the co-segmentation baseline~\cite{zhong2018_petctunet} with PET-CT patch inputs, which suggests that our CNN does not rely upon cropped data to learn. The overall lower Dice scores when compared to the baseline publications (which reported scores of about 80\%) is primarily attributed to the difficulty of automatically segmenting images with hilar (Stage II) and mediastinal (Stage III) nodal involvement, which is more challenging than segmenting tumors within the lung field. We note that other methods in the literature that are capable of segmenting tumors across the entire image are not automated and require human input to define the tumor bounding box or seed points for initial segmentation~\cite{song2013_petcttumsegsepgraph,ju2015_randwalkcutsegpetct}.

The co-segmentation baseline~\cite{zhong2018_petctunet} follows a similar approach to the MB fusion baseline; the individual modalities are processed separately and the predicted regions can then be merged. Fig.~\ref{fig:res_seg} shows that the CT image alone is insufficient to identify all the tumors within the image: the heterogeneous tumor within the lung is under-segmented while one of the two mediastinal tumors was not segmented. The PET segmentation detects three tumors but only the regions with the highest SUV have been detected leading to under-segmentation. The integration of the PET and CT result obtains an appropriate ROI for the primary tumor, but the disease outside the lung fields are not well-defined. The variational baseline~\cite{li2019_dlpetctseg} relies upon the initial segmentation of the CT image with refinement of the boundaries via PET. However, the CT image is insufficient for the mediastinal tumour segmentation, resulting in poor overall segmentation performance.

In comparison to the baselines, our CNN was able to identify and segment tumors consistently from the full PET-CT image slice. This outcome was due to its ability to consider PET and CT information in an integrated manner, with the fusion maps balancing the modality-specific information. The results indicate that our method is capable of segmenting tumors across different anatomical locations without prior cropping of patches around the tumor region.

\subsection{Fusion Map Analysis}
\label{sec:discuss:fusion}

The manner of feature fusion is a key difference in our CNN versus the baseline fusion CNNs. Our CNN derives a fusion map for each image that is explicitly multiplied across the feature maps of the different modalities (see Equation~\ref{eq:fusion}), thereby acting as feature weights. As such, our method can potentially derive different fusion maps for different input PET-CT images, prioritizing different characteristics at different locations. In contrast, all the baseline fusion CNNs use the convolution operation to merge the different modalities without any consideration of the spatial relevance of the underlying data. Our CNN also involves such convolutions but they occur \emph{after} the prioritization of information by multiplication with the fusion map.

The fusion maps shown in Fig.~\ref{fig:res_fusion} demonstrate that our co-learning CNN can derive spatially varying fusion maps for images that contain multiple structures that have different fusion requirements. Our CNN does not need to divide the problem into distinct tasks for each ROI but rather can derive the relevant fusion information in an end-to-end manner. For example, the 7th CT fusion channel (row 1, column 7) and the 13th CT fusion channel (row 2, column 5) emphasize the lung fields relative to the area containing the tumor. We suggest the co-learning unit has produced these specific fusion channels because (in combination with other channels) they contain information to distinguish the lung fields from any tumors they may contain. It is well-established that the lung fields can be identified using CT data alone~\cite{hu2001_lungseg}, and it was expected that the co-learning CNN would operate in a similar fashion. The fusion maps automatically derived by our co-learning CNN prioritize the CT data for the lung field ROIs, consistent with this expectation. Similar patterns for lung fields are noticed in the fusion maps of other PET-CT images. 

While it may appear that several channels in the fusion map are redundant (similar in appearance to other channels), this is merely a visualization issue caused by normalizing 32-bit floating point greyscale images for display within the paper. As shown in Fig.~S4 in the Supplementary Materials, PET fusion channels 33 to 37 (row 13, columns 1 to 5) appear visually similar but closer examination of the distribution of fusion weights within the images indicates that each channel prioritizes information in subtly different ways. Section~SIII in the Supplementary Materials contains a detailed example showing the differences in these  visually similar fusion channels and their impact in the analysis of heterogeneous tumors, which is an important clinical application. We suggest that the capacity of our co-learning CNN to derive these subtly different fusion weights enables more precise integration of the complementary information in each modality when compared with uniform fusion (see Fig.~\ref{fig:res_fusion_compare}).

\subsection{Directions for Further Research}
\label{sec:discuss:future}

In our experiments, we compared our co-learning concept for fusion to other fusion approaches. To focus mainly on the differences in the approach to fusion, we built variant baseline CNNs that were similar to our CNN's architecture but that implemented fusion using conventional techniques. This was done so that the main difference between the baselines and our CNN was the presence of our co-learning component, limiting the number of architectural differences. It also meant that we could use similar hyperparameters for fairer experimental comparisons. Our findings indicate that the addition of the co-learning component improved the final results and as such we suggest that other CNNs may also see improvements if they were to follow a similar conceptual approach for feature fusion; we have left this for future research.

Similarly, we suggest our co-learning CNN could also be extended or adapted to be better optimized for different datasets and applications. Such extensions could be the inclusion of improved encoders that go beyond stacked CNNs by borrowing designs from Residual~\cite{he2016_resnet}, Inception~\cite{szegedy2017_inception}, or other newer CNN architectures; enhanced application-specific encoders would better optimize the feature extraction for different applications. Similarly, the co-learning unit could also be similarly adapted with multiple stacked convolutions to derive fusion maps with even finer details. Finally, it is expected that the final blocks of the reconstruction component will be redesigned for different applications, e.g., such as by using fully connected layers or global average pooling~\cite{selvaraju2017_gradcamglobavgpool} for classification applications. In addition, our approach could be adapted with multi-resolution fusion techniques~\cite{ghassemian2016_multiresfusreview} to enhance its capacity to extract and integrate information at different scales. To compute the fusion maps, our CNN leverages some multi-scale information: each encoder block in the architecture processes the downscaled feature maps from the previous encoder block (see Fig.~\ref{fig:arch}). This is done with max pooling to ensure that the most prominent aspects of the feature maps influence the fusion map computation and the extraction of features from the following encoder blocks. Integration of a multi-resolution fusion technique such as RefineNet~\cite{lin2017_refinenet}, which uses a separate path for each scale or resolution, could further enhance the ability to analyze and refine ROIs with subtle boundaries. We will examine some of these adaptations in our future research.

Our method required selection of the extent of the thorax so its transaxial slices could form the CNN input. This is currently a manual process in our dataset construction to ensure that the correct body region could be analyzed. Automation of this process could result in a performance decrease if slices outside this body region were selected. Hence, an interesting area for future research would be to automatically classify the slices of whole body PET-CT so that they could be correctly passed to the CNN for analysis.

Our current CNN design also requires images in the same coordinate space so that the spatially varying fusion can be computed. However, the  PET and CT images originally have different resolutions (see Section~\ref{sec:methods:materials}) and so we rescaled both modalities to $256\times256$ pixels. We note that downscaling the CT data may cause subtle details near tissue boundaries (bronchus or chest wall) to be lost while upscaling the PET may introduce further noise due to the interpolation of the pixel data. We elected to rescale both modalities to a resolution that was in-between the original resolutions to limit the possible the distortions for each modality. The distortions due to data rescaling could possibly be avoided by performing spatially varying fusion across images with different resolutions, but this is a significant challenge that is beyond the scope of this research and hence we leave it to future work.

Our experiments used a dataset of 50 PET-CT images, which is of a similar scale to other studies that have used multi-modality image data~\cite{xu2018_petctbone,zhong2018_petctunet,zhang2015_neuromm,zhong2019_simultaneouscosegpetct}. To reduce the likelihood of overfitting, we trained our co-learning CNN using standard data augmentation techniques (see Section~\ref{sec:methods:train}). Furthermore, to reduce the likelihood of biased results, we used a 5-fold cross-validation analysis protocol, which is a valid compromise in the absence of larger datasets. We note that the use of larger datasets would provide greater opportunities for our co-learning CNN to learn the appropriate fusion characteristics of the underlying data.

In addition, we used greyscale inputs for all experiments rather than use color lookup tables (CLUTs) for PET. CLUTs are sometimes used to enhance the appearance of functional information, particularly in image visualization. Our experimental aim was to focus on how the information from each modality was prioritized and the colorization of PET may have biased the functional information. However, we acknowledge that color information may provide additional visual features and we will explore this in a future study.

Our experiments focused on verifying our CNN’s capabilities in co-learning the fusion on lung cancer where the fusion requirements are well-understood. There are other problem domains (e.g., whole body disease such as lymphoma or metastatic cancer) that contain a wider variety of anatomical structures and heterogeneous disease. These problem domains will require ROI-specific fusion, which may not be known \emph{a priori}. The ability of our co-learning CNN to derive fusion maps for different structures suggest that it could potentially be applied to generate ROI-specific fusion maps for these problems. Furthermore, we used a supervised learning approach and hence the current CNN design and the trained model was focused on application to a specific known disease. This is consistent to other CNN methods that are tuned for specific diseases or tasks ~\cite{anthimopoulos2016_cnnild,christ2016_liverlesionfcncrf}. We suggest that given sufficient training data, there would be no barrier to developing co-learning CNNs for other specific tasks or for deriving CNNs capable of deriving fusion maps across a range of related applications. These are also avenues for future work.

\section{Conclusion}
\label{sec:conc}

We presented a new supervised CNN for fusing complementary information from multi-modality images. Our CNN leveraged modality-specific features to derive a spatially varying fusion map that quantified the importance of each modality's features across different spatial locations. Our findings from region detection and segmentation experiments on PET-CT lung cancer images demonstrated that our method was a significant ($p < 0.05$) improvement upon several baseline CNN-based methods for multi-modality image analysis. We suggest that our conceptual approach of having a specific CNN architectural component to derive explicit fusion maps could be a useful technique for medical image analysis applications that require considering complementary information from different image modalities, e.g., PET-CT and PET-MR.

\bibliographystyle{IEEEtran}
\bibliography{bibtex/bib/refs}


\onecolumn
\newpage

\setcounter{section}{0}
\renewcommand*{\theHsection}{chY.\the\value{section}}
\setcounter{figure}{0}
\renewcommand*{\theHfigure}{chY.\the\value{figure}}

\newcommand\largesection{%
  \titleformat*{\subsection}{\normalfont\Huge\filcenter}
}

\makeatletter 
\renewcommand{\thesection}{S\@Roman\c@section}
\makeatother
\makeatletter 
\renewcommand{\thefigure}{S\@arabic\c@figure}
\makeatother

\largesection
\subsection*{Supplementary Materials}
\bigskip

\section{Verification of Parameter Settings}
\label{sec:parameter}

\begin{figure}[!h]
	\centering
    \includegraphics[width=1\columnwidth,keepaspectratio]{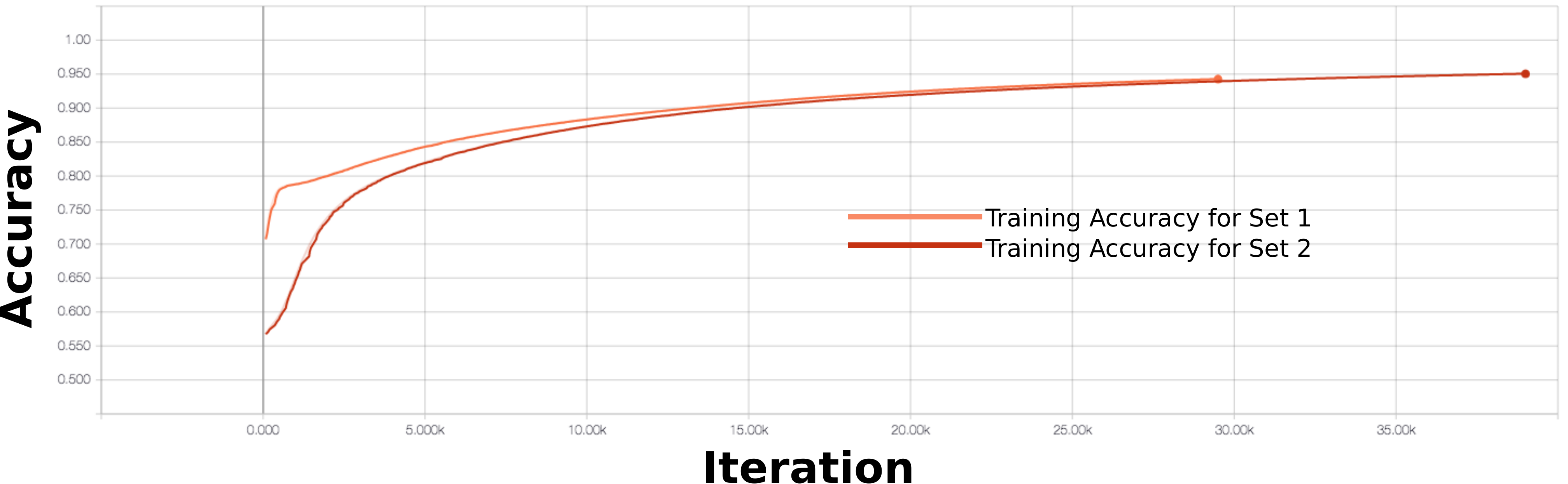}
    \caption{Tensorboard output for the training accuracy in both set.}
    \label{fig:train}
\end{figure}

\begin{figure}[!h]
	\centering
    \includegraphics[width=1\columnwidth,keepaspectratio]{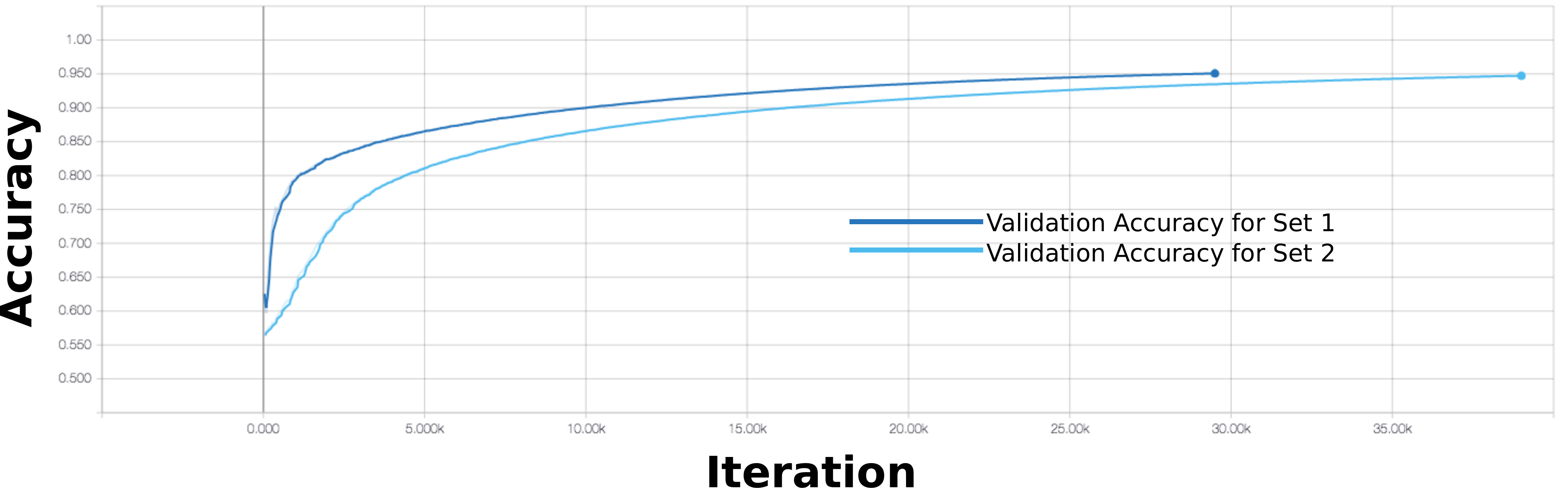}
    \caption{Tensorboard output for the validation accuracy in both set.}
    \label{fig:val}
\end{figure}

We used a cross-validation approach to verify that the architecture and training parameters were appropriate for our dataset. For this verification we used the 40 studies in Fold~1 of the training dataset (see Section~{III-A} in the main body of the paper), which we divided into two sets of data, each comprising the slices from 20 studies. Slices from the same study were restricted to the same set, i.e., no slices from one study appeared in both sets. As the studies in the sets were randomly selected, each set contained a different number of individual slices. For this reason, the second set had a larger number of iterations compared to the first set over the same number of epochs. Fig.~\ref{fig:train} shows the Tensorboard logs of the training accuracy across both sets, recorded every 20 iterations. Similarly, Fig.~\ref{fig:val} shows the Tensorboard logs of the validation accuracy across both sets, recorded every 20 iterations. The figures show very similar levels of training and validation accuracy across both sets (0.943 training, 0.951 validation for Set~1; 0.951 training, 0.947 validation for Set~2), despite the difference in set size, indicating that the parameters of our architecture resulted in stable learning.

\newpage

\section{Results: Larger Images}
\label{sec:reshighres}

\begin{figure}[!h]
	\centering 		     	
    \includegraphics[width=0.72\textwidth,keepaspectratio]{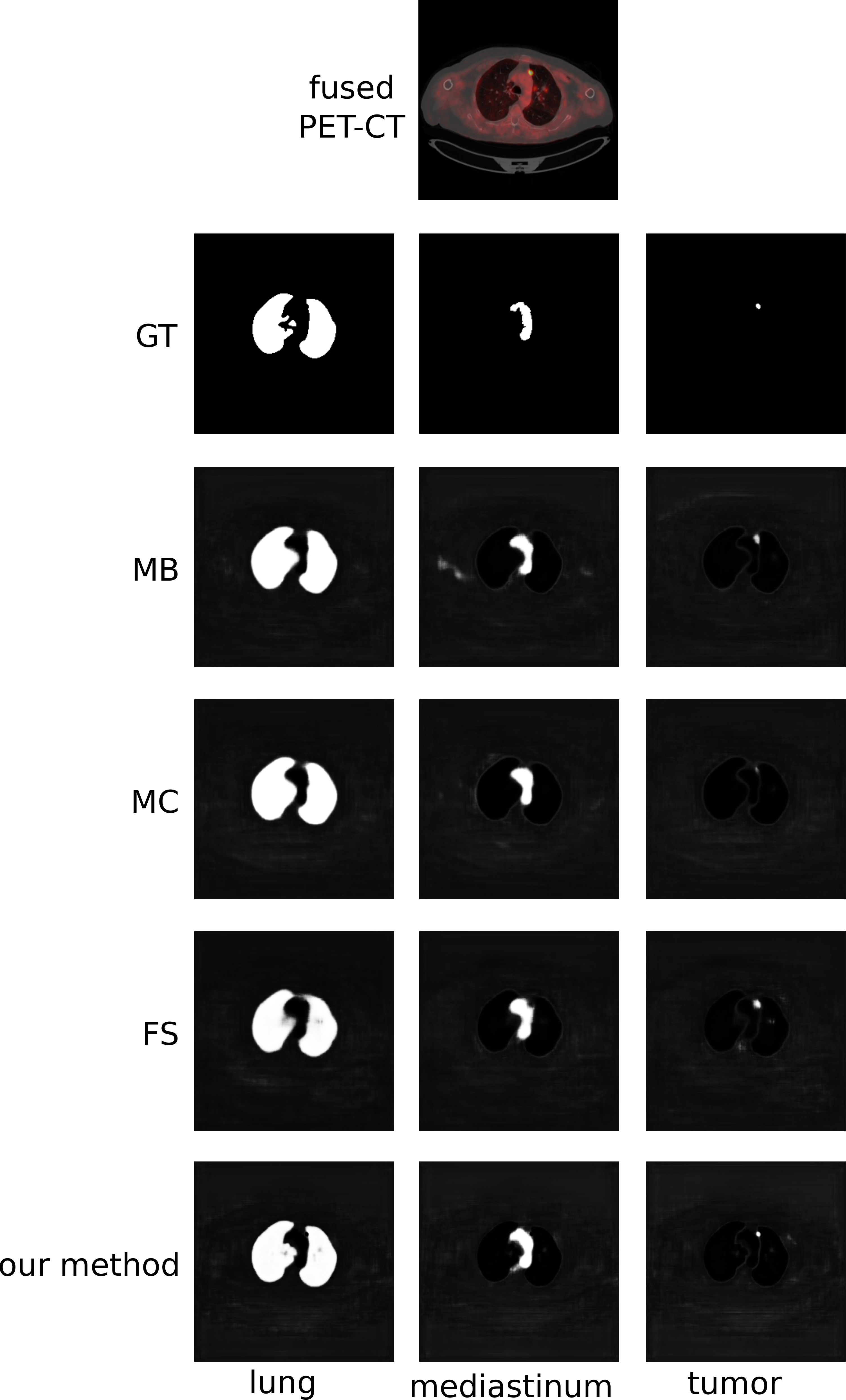}
    \caption{Larger rendering of the visual results in the main text.}
    \label{fig:highresdetect}
\end{figure}

Figure~\ref{fig:highresdetect} is a larger version of Figure~5 from the main text to show the results in greater detail. 

\newpage

\begin{figure}[!h]
	\centering 		     	
    \includegraphics[width=\textwidth,keepaspectratio]{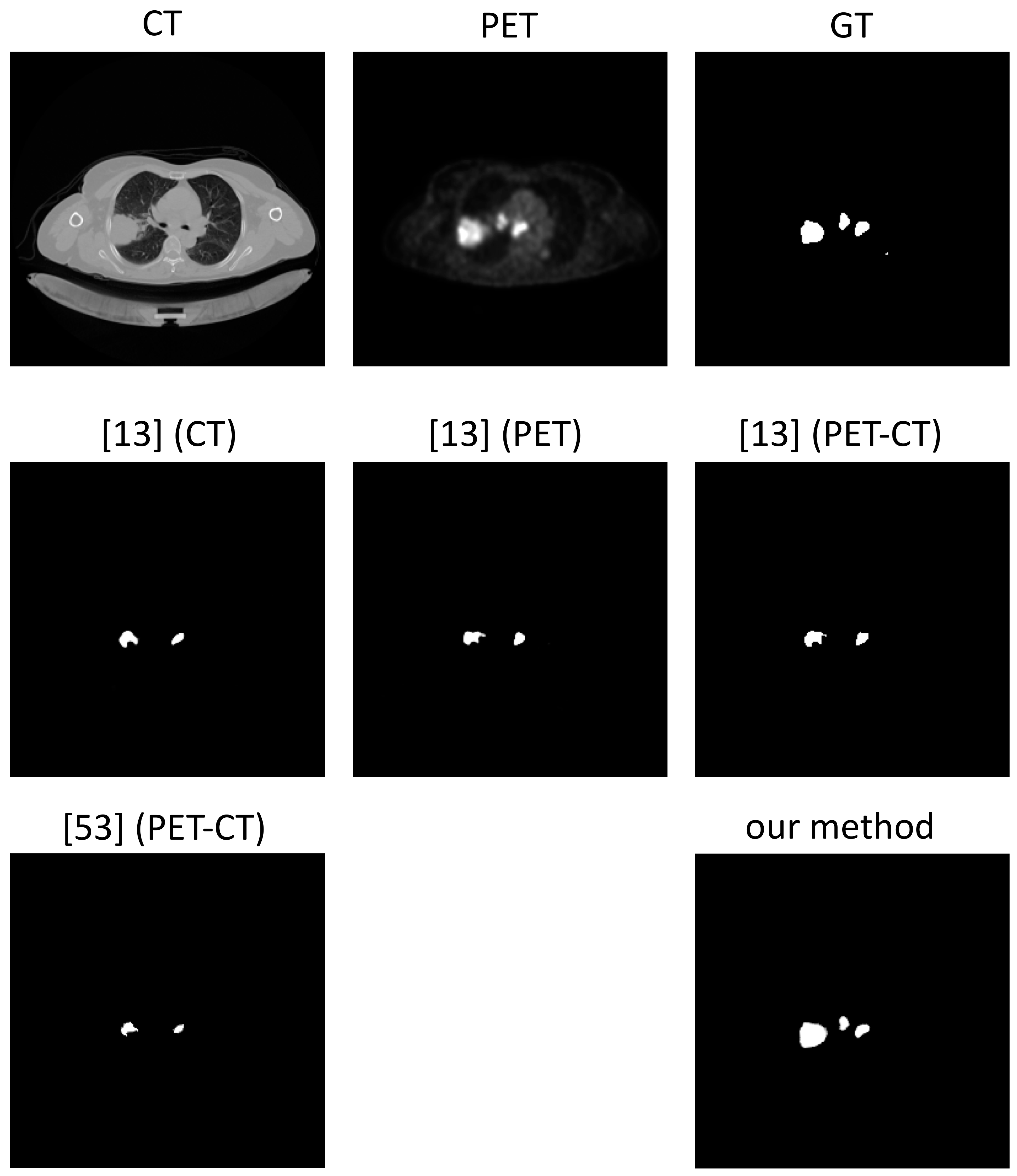}
    \caption{Larger rendering of the visual results in the main text.}
    \label{fig:highresseg}
\end{figure}

Figure~\ref{fig:highresseg} is a larger and rearranged version of Figure~6 from the main text to show the results in greater detail. 

\newpage

\begin{figure}[!h]
	\centering 		     	\includegraphics[width=0.7\textwidth,keepaspectratio]{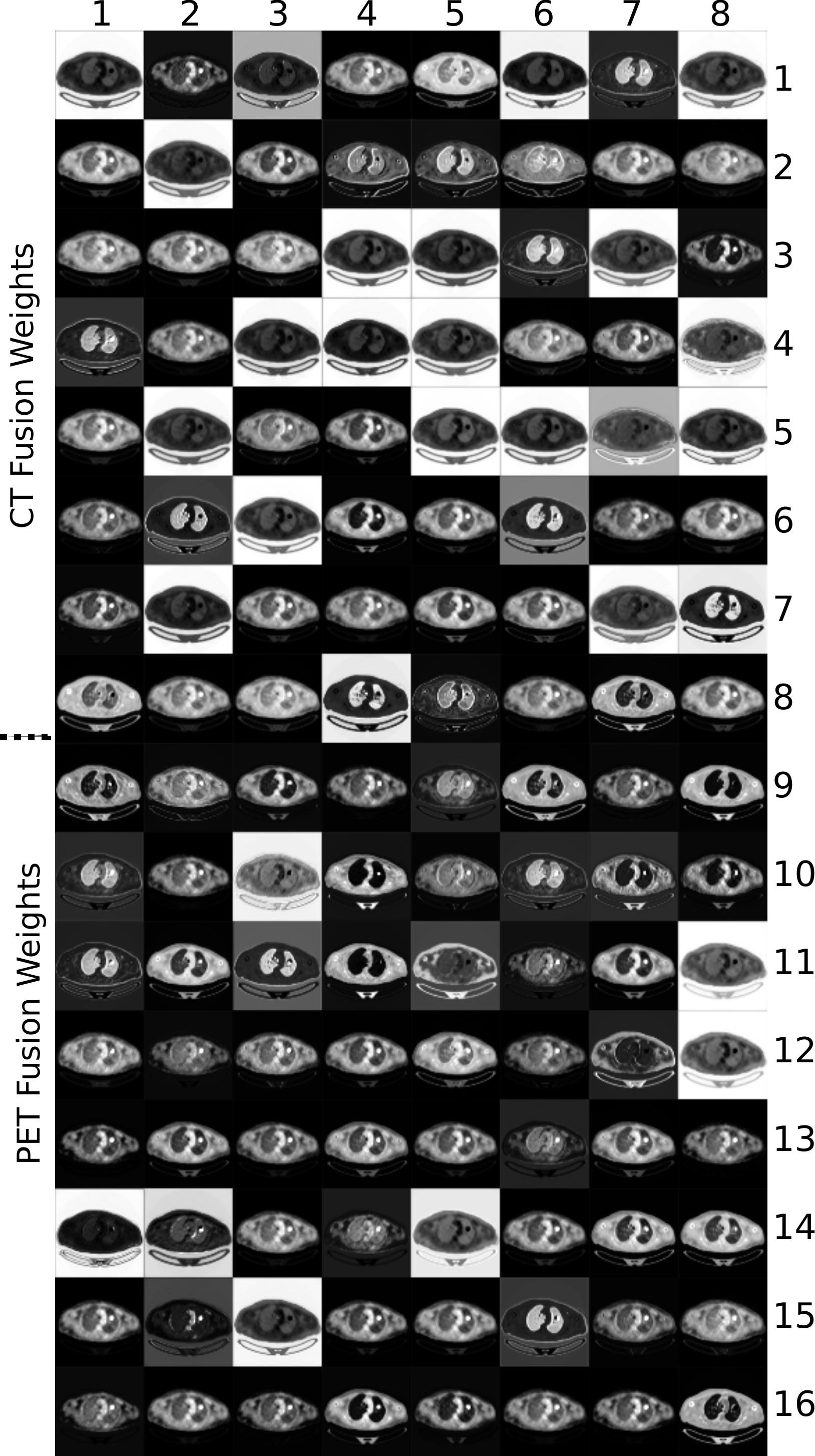}
    \caption{Larger rendering of the fusion maps in the main text.}
    \label{fig:highresfuse}
\end{figure}

Figure~\ref{fig:highresfuse} is a larger version of Figure~8 from the main text, showing all 128 fusion maps from the first co-learning unit. 

\newpage

\begin{figure}[!h]
	\centering 		     	\includegraphics[width=0.45\textwidth,keepaspectratio]{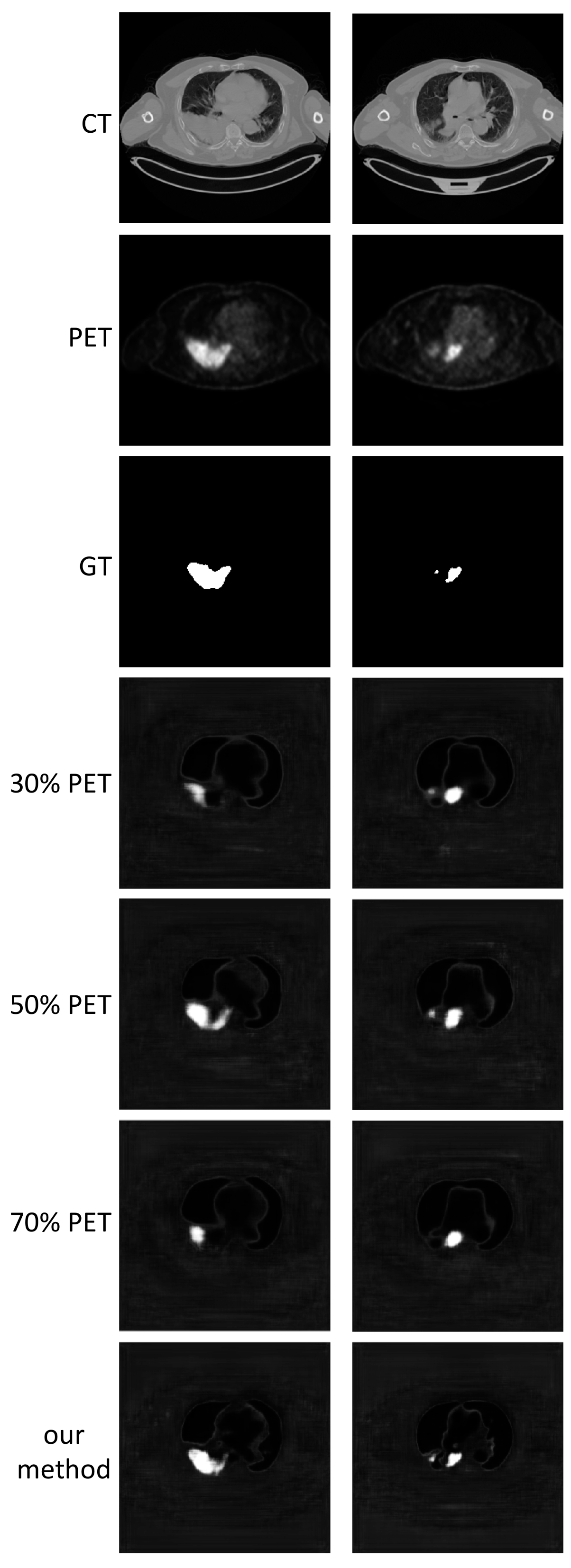}
    \caption{Larger rendering of the comparative results of different fusion methods in the main text.}
    \label{fig:highfusecomp}
\end{figure}

Figure~\ref{fig:highfusecomp} is a larger version of Figure~7 from the main text to show the results in greater detail. 

\newpage

\section{Fusion Maps}
\label{sec:fusion}

\begin{figure}[!h]
	\centering
    \includegraphics[height=0.8\textheight,keepaspectratio]{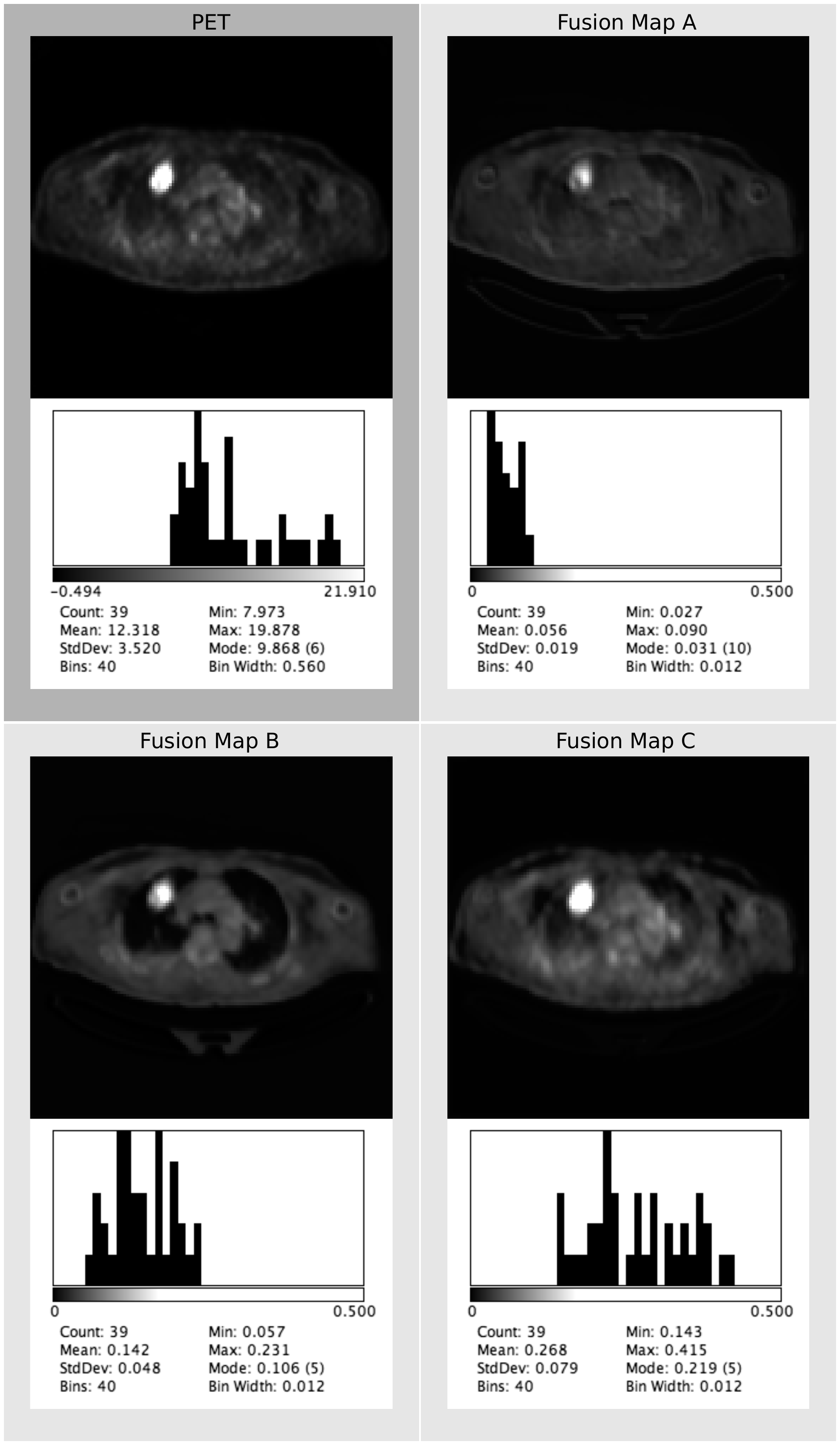}
    \caption{Examination of tumor in the PET image against the fusion map characteristics within the tumor region.}
    \label{fig:fusionmaps}
\end{figure}

Figure~\ref{fig:fusionmaps} is an analysis of the fusion maps generated by our co-learning units. In this example, we examine the distribution of pixels within the tumor region within the PET image and three channels of the generated fusion map.

The tumor region has high intensity compared to the other parts of the image and as such it is difficult to visually ascertain the differences among the fusion maps. As such, we used the ground truth tumor region and calculated the intensity histogram for the pixels within the tumor in the PET image and in the fusion maps; these histograms are also shown in Figure~\ref{fig:fusionmaps}.

The histogram of the PET image shows that the tumor is heterogeneous, with a maximum SUV of approximately 20. The mode of the tumor pixels are at the tail end of the distribution (approximate SUV of 10), below the mean SUV of 12. Overall the distribution of the tumor is skewed towards the lower SUV values.

The fusion maps all have distributions that are different to the tumor's original SUV distribution and are also distinct from each other. The tumor region within Fusion Map~A has a relatively homogeneous distribution where the weights are clustered: the minimum and maximum fusion weights (coefficients) are within two standard deviations of the mean. A potential interpretation is that this fusion map differentiates the tumor region from the surrounding non-tumor areas, which have lower intensity as seen in the image. In contrast, Fusion Maps~B and~C have histograms showing heterogeneous distributions; neither of these match the distribution pattern of the original tumor's SUV distribution, implying that they are each prioritizing different aspects of the tumor region.

\section{Dataset Creation Process}
\label{sec:datasetprocess}

\begin{figure}[!h]
	\centering
    \includegraphics[width=\textwidth,keepaspectratio]{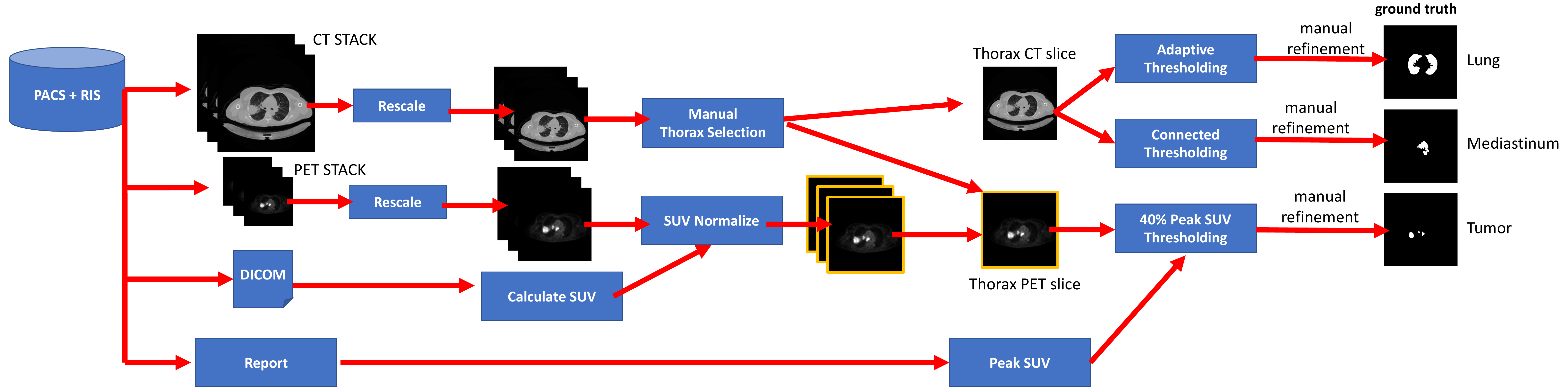}
    \caption{Our data creation process.}
    \label{fig:data}
\end{figure}

The dataset creation process is described below. A diagrammatic form of the image preprocessing and ground truth creation is illustrated in Fig.~\ref{fig:data}.

\begin{enumerate}
    \item Select 50 PET-CT lung cancer studies and their corresponding clinical reports from the hospital Picture Archiving and Communication System (PACS) and Radiology Information System (RIS).
    \item For every study, calculate the standardized uptake value (SUV) coefficients from the information in the DICOM headers.
    \item Using the DICOM information on pixel spacing, rescale the CT and PET image volumes to the same coordinate space ($256\times256$ pixels in the x-y axes for our research).
    \item Transform each PET volume to an SUV normalized form using the study-specific SUV coefficients calculated in Step 2.
    \item For each study, note the peak (maximum) tumor SUV listed in the corresponding clinical report.
    \item For each study, use the transaxial plane of the CT volume to identify the start and end of the thorax region.
    \item For each study, apply adaptive thresholding to the CT thorax subvolume to extract the lung regions. The thresholding parameters may require manual adjustment. This is the initial mask that may require manual refinement to form the ground truth for the `lungs' class.
    \item For each study, apply connected thresholding to the CT thorax subvolume to extract the mediastinal regions. The thresholding parameters may require manual adjustment. This is the initial mask that may require manual refinement to form the ground truth for the  `mediastinum' class.
    \item For each study, apply connected thresholding to the SUV-normalized PET thorax subvolume; the lower bound is 40\% of the peak SUV calculated in Step 4, while the upper bound is the peak SUV value. If required, manually adjust the masks of the tumor regions, using the confirmed clinical reports to reference  anatomical location and invasion into adjacent structures. This is the ground truth for the `tumor' class.
    \item Randomly divide the studies into 5 different folds for training and testing. Each training fold comprises the images from 40 studies, while each test fold comprises the images from the remaining 10 studies.
    \item Using the ground truth for each study, select the image slices that contain regions from all classes. This guarantees an even distribution for each class.
\end{enumerate}

\section{Source Code}
\label{sec:sourcecode}

The refactored source code and documentation for our CNN can be found on \url{https://github.com/ashnilkumar/colearn}.

\end{document}